%% file: main.tex
\documentclass{article}
\usepackage{style_1,times}

\usepackage{microtype}
\usepackage{graphicx}
\usepackage{booktabs}
\usepackage{hyperref}
\usepackage{url}

\usepackage{amsmath}
\usepackage{amssymb}
\usepackage{mathtools}
\usepackage[capitalize,noabbrev]{cleveref}
\crefname{appendix}{Appendix}{Appendices}
\Crefname{appendix}{Appendix}{Appendices}

\newcommand{\safefig}[2]{%
  \IfFileExists{#2.pdf}{\includegraphics[width=#1]{#2.pdf}}{%
  \IfFileExists{#2.png}{\includegraphics[width=#1]{#2.png}}{%
  \fbox{\parbox[c][3cm][c]{\dimexpr#1-2\fboxsep-2\fboxrule\relax}{\centering\textbf{MISSING FIGURE:} \texttt{\detokenize{#2.pdf}}}}}}}

\newcommand{\pagefig}[3]{
\begin{figure}[p]
\centering
\IfFileExists{#1.pdf}{\includegraphics[width=\linewidth,height=0.86\textheight,keepaspectratio]{#1.pdf}}{%
  \fbox{\parbox[c][3cm][c]{0.8\linewidth}{\centering\textbf{MISSING FIGURE:} \texttt{\detokenize{#1.pdf}}}}}
\caption{#2}
\label{#3}
\end{figure}}

\newcommand{\nullfig}[2]{\IfFileExists{#2.pdf}{\includegraphics[#1]{#2.pdf}}{\fbox{MISSING FIGURE: \detokenize{#2}}}}

\newcommand{\nullpage}[2]{
\begin{figure}[p]
\centering
\nullfig{width=\linewidth,height=0.70\textheight,keepaspectratio}{null_#1_all_layers}\\[4pt]
\nullfig{width=0.9\linewidth,height=0.17\textheight,keepaspectratio}{null_#1_first_to_last}
\caption{Null controls for #2 on CIFAR-10: as \cref{fig:null_rate} with every layer (top) and as \cref{fig:null_f2l} (bottom).}
\label{fig:null_#1}
\end{figure}}

\newcommand{\nullpageds}[4]{%
\begin{figure}[p]
\centering
\nullfig{width=\linewidth,height=0.70\textheight,keepaspectratio}{null_#1_#2_all_layers}\\[4pt]
\nullfig{width=0.9\linewidth,height=0.17\textheight,keepaspectratio}{null_#1_#2_first_to_last}
\caption{Null controls for #3 on #4: as \cref{fig:null_rate} with every layer (top) and as \cref{fig:null_f2l} (bottom).}
\label{fig:null_#1_#2}
\end{figure}}

\title{Rescuing Points Trapped Inside Incorrect\\Decision Regions: A target for Evaluating and Improving Forward-Forward Learning}

\makeatletter
\def\@maketitle{\vbox{\hsize\textwidth
{\centering\LARGE\sc \@title\par}
\vskip 0.2in
{\@author}
\vskip 0.3in minus 0.1in}}
\makeatother
\renewcommand{\headrulewidth}{0pt}

\newcommand{\authorcell}[1]{\begin{tabular}[t]{@{}c@{}}#1\end{tabular}}
\author{%
\begin{tabular*}{\textwidth}{@{\extracolsep{\fill}}ccc@{}}
\authorcell{\textbf{Alireza Rahmansetayesh}\\
Department of Computer Science\\
University of California, Irvine\\
Irvine, CA, USA\\
\texttt{arahmans@uci.edu}}
&
\authorcell{\textbf{Amirhossein Yousefiramandi}\\
Independent Researcher\\
\texttt{amir.usefi75@gmail.com}}
&
\authorcell{\textbf{Maxwell Spiridonov}\\
Department of Computer Science\\
University of California, Irvine\\
Irvine, CA, USA\\
\texttt{mspirido@uci.edu}}
\\[1.5ex]
\authorcell{\textbf{Sumit Raut}\\
Department of Computer Science\\
University of California, Irvine\\
Irvine, CA, USA\\
\texttt{scraut@uci.edu}}
&
\authorcell{\textbf{Kent Ziti}\\
Department of Computer Science\\
University of California, Irvine\\
Irvine, CA, USA\\
\texttt{kziti@uci.edu}}
&
\authorcell{\textbf{Pierre Baldi}\\
Department of Computer Science\\
University of California, Irvine\\
Irvine, CA, USA\\
\texttt{pfbaldi@ics.uci.edu}}
\end{tabular*}}
\date{}

\iclrfinalcopy

\begin{document}

\maketitle

\begin{abstract}
Forward-Forward learning replaces end-to-end backpropagation with local, layer-wise training objectives. Its variants are evaluated primarily by classification accuracy, often measured with linear probes trained on their representations. We show that this evaluation can over-credit the learning method: a linear probe on the representations of an untrained convolutional neural network already reaches accuracy well above chance. Moreover, a single accuracy value does not reveal which datapoints become correctly classified after training and which remain misclassified. To examine what learning contributes beyond the random, untrained baseline, we present an analysis methodology built on the Boundary Proximity Index (BPI), a multiclass margin normalized by a class-centroid reference. We apply our analysis to seven Forward-Forward variants and related layer-local methods and find that initially misclassified points near the probe’s decision boundary generally have higher rescue rates than those farther inside incorrect decision regions. Rescue rates vary substantially across methods, with some correcting a substantial fraction of these deeply embedded points.  Together, these analyses offer a way to evaluate representation learning beyond accuracy and identify the rescue of points far inside incorrect decision regions as a target for improving local learning methods.
\end{abstract}

\section{Introduction}
\label{sec:intro}

Backpropagation (BP)~\citep{rumelhart1986learning,lecun1998convolutional,baldi2021deep} is the dominant way of training deep networks, but its global error signals, weight transport and backward pass raise concerns about biological plausibility and computational cost~\citep{crick1989recent,baldi2017learning}. This has motivated learning methods that are forward-only and layer-local. A prominent one is the Forward-Forward algorithm~\citep{hinton2022forward}, which has been extended to convolutional networks and deeper architectures in many ways~\citep{scodellaro2025training,lorberbom2024layer,lee2023symba,lee2025vff,dooms2024the,chen2025self,sun2025deeperforward,yao2026hclff,sarode2026hyperspherical,gong2025asge}. These methods are almost always compared through test accuracy of the layers of the network, typically read out by a linear classifier on the learned representations.

Final accuracy is a coarse summary, and it can mislead in two ways. First, convolutional architectures are strong feature extractors even with random weights~\citep{jarrett2009best,saxe2011random,alain2018understanding}, so part of the accuracy credited to a learning method may be contributed by the architecture. Second, accuracy does not reveal \emph{which} datapoints a method helps, which ones remain misclassified, and what makes them hard to correct.

To address these questions, we study each datapoint's position relative to the decision boundaries of a linear probe on the learned representations, across layers and before and after training. Our tool is a \emph{pointwise} measure derived from the probe, the \emph{Boundary Proximity Index} (BPI): the ordinary multiclass margin~\citep{crammer2001algorithmic,koltchinskii2002empirical}, normalized (in units set by the class centroids) to give a dimensionless reference across layers of different dimensionalities. The sign of BPI states on which side of the probe's decision boundary a datapoint lies, and its magnitude how far from it. Whereas an aggregate score answers, ``Did the representation improve?'', this pointwise measure answers, ``How does the likelihood that a datapoint improves depend on its initial BPI?''

We build a methodology around BPI with two parts. \emph{Across training}: at every layer, we compare the classification outcome of each datapoint under a linear probe on the untrained network and on the trained network of the same architecture and seed. A datapoint misclassified before training and correctly classified after is \emph{rescued}, and one correctly classified before and misclassified after is \emph{broken}; we study how rescue and breakage rates depend on the datapoint's BPI in the untrained representation before learning. \emph{Across depth}: we follow a fully trained network from one layer to the next, and from the first to the deepest layer, and measure the rescue and breakage of datapoints as a function of their BPI at the layer's input. Since all earlier layers are already trained, this shows how much a layer adds to linear separability beyond the layers before it. This methodology is method-agnostic and applies to any learning method whose representations are read out linearly and are partially linearly separable.

Our analysis, and BPI itself, depend on linear probes trained on the representations, which adds uncertainty. For probes to be informative, their training must be stable and reach accuracies well above chance, which requires some linear structure in the representation. We observe that probes are well above chance in every layer (the exception is the raw-pixel probe on Tiny ImageNet and ImageNet-1K, Appendix~\ref{app:datasets}), and refitting a probe with a different seed changes the classification of only 0.4 to 3.1\% of the test datapoints (Appendix~\ref{app:null3}). Partial linear separability of untrained representations is also well documented~\citep{jarrett2009best,saxe2011random,rahimi2007random}. We also performed three null-control experiments on untrained networks, each changing one ingredient at a time (\cref{sec:null}). \emph{Null~1} changes nothing but the probe: two linear probes with different seeds are fitted on the same untrained features, so that any rescue is probe noise; \emph{Null~2} replaces the trained network by a second untrained network with a different random initialization; \emph{Null~3} applies the across-depth analysis to an untrained network. In all three, datapoints are binned by BPI under the first representation or probe, as for the trained networks. These controls justify the use of BPI: probe noise changes the classification of almost no datapoint far from the boundary (Null~1), and Nulls~2 and~3 provide the baseline without learning against which the contribution of learning can be measured.

We study seven layer-local learning methods: SCFF~\citep{chen2025self}, Trifecta~\citep{dooms2024the}, DeeperForward~\citep{sun2025deeperforward}, HCL-FF~\citep{yao2026hclff}, HFF~\citep{sarode2026hyperspherical}, Mono-Forward~\citep{gong2025monoforward} and ASGE~\citep{gong2025asge}, mainly on CIFAR-10, CIFAR-100 and Tiny ImageNet. Local learning methods remain behind backpropagation in efficiency and accuracy, so most original articles report results only on such datasets. Some also report results on ImageNet-1K, but with accuracies well below the state of the art; for example, HFF~\citep{sarode2026hyperspherical} reports $25.70\%$ top-1 with VGG16. Its code is not public and many hyperparameters are not given; in our best-effort re-implementation, the method's own read-out reaches $7.7\%$ top-1, and linear probes on the learned features reach up to $15\%$. Because our HFF re-implementation falls substantially below the reported accuracy, its results characterize this trained instance rather than the performance attainable by HFF. We include it to illustrate the application of the analysis to ImageNet-1K, not as evidence that a faithfully reproduced HFF model has the same limitations. The accuracies in this paper should not be compared with those of state-of-the-art conventional work. The main text presents CIFAR-10; the corresponding figures for the other datasets, for the methods run on them, are in Appendix~\ref{app:datasets}.

The findings are qualitative and, on CIFAR-10, CIFAR-100 and Tiny ImageNet, consistent across the methods run on each dataset (Appendix~\ref{app:datasets}); the ImageNet-1K analysis is exploratory. (i)~Untrained networks supply a large share of final accuracy; the share differs widely between methods, and for some methods training improves the probe accuracy only slightly. (ii)~Whether training rescues a misclassified datapoint depends strongly on where its untrained representation lies relative to the decision boundary: points near the boundary are rescued more often than points deep on the wrong side. (iii)~The same holds inside a trained network: whether a layer rescues a datapoint that the previous layer misclassified depends on its BPI at that layer's input, and an additional layer generally rescues points closer to the decision boundary than points with very negative BPI. Learning methods differ in how much they rescue, and stronger methods rescue more points at every BPI level. However, we still see that the more negative the BPI, the lower the rescue rate. (iv)~Untrained networks show the same rescue-BPI relationship at a lower level: a second random initialization, or an untrained layer, also rescues points near the boundary and almost none far from it.

Our contribution is an evaluation methodology combining architecture-matched untrained baselines, layer-wise training ablations, BPI-conditioned rescue and breakage rates, and null controls. The monotone relationship between rescue and BPI is not specific to local learning and is not itself evidence of successful training. The informative differences are the magnitude of rescue achieved by each method, its relationship to the null controls, and the additional benefit obtained from training deeper layers. 

\section{Results}
\label{sec:results}

\subsection{Untrained CNNs Provide a Strong Baseline for Measuring the Benefit of Training}
\label{sec:untrained}

\begin{figure}[t]
\begin{center}
\safefig{\linewidth}{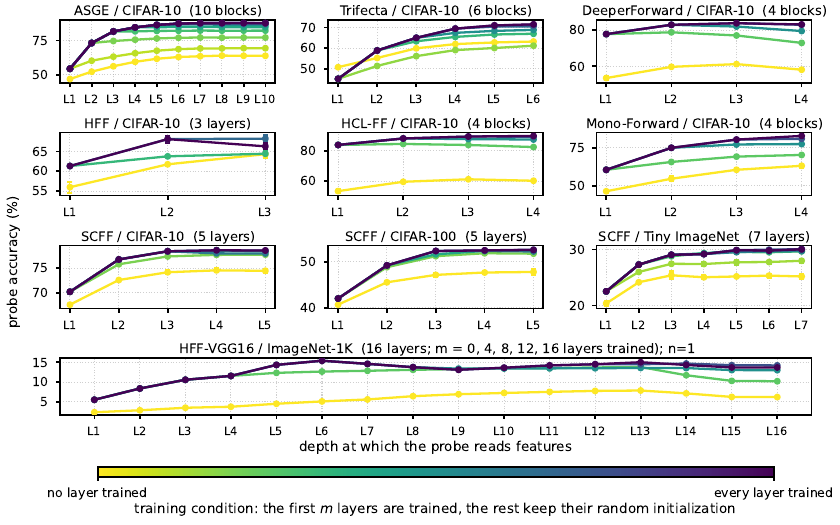}
\caption{Layer-wise ablation of training. Each panel is one method and dataset (mean and standard deviation over three seeds). A curve is the test accuracy of a linear probe against the depth at which it reads features, for one training condition: the color encodes how many of the first layers are trained, from none (yellow, the untrained network) to all (dark purple, the usual reported result), the rest keeping their random initialization. The gap between the yellow and the dark-purple curve is the improvement from training under this probe protocol. The SCFF networks include two appended layers (Appendix~\ref{app:setup}). The ImageNet-1K panel (HFF-VGG16, trained by us) shows five of the 17 conditions (0, 4, 8, 12 and 16 layers trained).}
\label{fig:ablation}
\end{center}
\end{figure}

Random convolutional networks are known to produce features on which a linear classifier achieves accuracy well above chance~\citep{jarrett2009best,alain2018understanding}, so a learning method should be judged against the same architecture with random weights. To measure the improvement from training under a fixed probe protocol, we ablate training layer by layer (\cref{fig:ablation}): we train only the first $m$ layers, leave the others at their random initialization, and read the representation at every depth $\ell$ with a linear probe. For every method, this representation is the concatenation of the pooled, per-sample standardized activations of layers $1,\dots,\ell$ (Appendix~\ref{app:setup}). The protocol combines information across layers, as several methods' original readouts do, but uses a common fitted linear probe rather than reproducing each method's native readout; Trifecta and HFF, whose readouts use the last layer only, are analyzed the same way. We ran this ablation for ASGE, Trifecta, DeeperForward, HFF, HCL-FF and Mono-Forward on CIFAR-10, for SCFF on all three datasets, and for HFF-VGG16 on ImageNet-1K (setup in Appendix~\ref{app:setup}).

On CIFAR-10, untrained feature extractors already support approximately 58–74\% linear-probe accuracy at the deepest cumulative representation. Training adds about 2 points for HFF, 4 to 5 for SCFF,  8 for Trifecta, and about 20 to 30 for Mono-Forward, ASGE, DeeperForward and HCL-FF.

\Cref{fig:ablation} shows that accuracy at a given layer depends on training in the preceding layers, which is important when assessing how effectively a method trains deeper layers. The contribution of training a layer should therefore be measured relative to a network in which the preceding layers are trained and that layer remains random, rather than one in which all layers are random. Let $a_\ell(m)$ denote the probe accuracy at depth $\ell$ when the first $m$ layers are trained. The contribution of training layer $\ell$ is then $a_\ell(\ell)-a_\ell(\ell{-}1)$, whereas the gap relative to the fully untrained network is $a_\ell(\ell)-a_\ell(0)$. For SCFF on CIFAR-10, training the first two layers already increases accuracy at the third layer, so training the third layer itself adds 0.0 percentage points, despite a 4.2-point gap relative to the fully untrained network.  This pattern holds across all ablated networks: from the fifth layer onward, training an additional layer contributes at most 1.6 points for ASGE, Trifecta, and SCFF on the three datasets. These contributions are small compared with the gaps of approximately 4 to 25 points relative to the fully untrained networks. Earlier layers can contribute substantially more; for example, the second layer of ASGE adds 12.8 points.

We therefore suggest that such an ablation be part of the evaluation of local-learning methods: the accuracy at each layer should be compared with the network in which the previous layers are trained and that layer is random, which isolates its effect from that of the layers before it.

\subsection{Rescue Probability Increases as the Initial Representation Approaches the Probe's Decision Boundary}
\label{sec:across_training}

Since the untrained representation already separates many datapoints, the question is which of the remaining errors learning can fix, and how it treats the datapoints that are already correct. Final accuracy cannot answer it: a misclassified datapoint near the decision boundary of a linear probe and one deep inside the region of a wrong class may need very different changes of the representation to be corrected. We therefore need a pointwise measure of position relative to the decision boundary, the Boundary Proximity Index (BPI), defined below.

\paragraph{Boundary Proximity Index.}
Let $\mathbf{r}\in\mathbb{R}^d$ be the representation of a datapoint with label $y$, and let $f_k(\mathbf{r})=\boldsymbol{\beta}_k^\top\mathbf{r}+b_k$ be the logits of a linear probe fitted on the training set. Let $\hat k=\arg\max_k f_k(\mathbf{r})$ be the predicted class and $k^\star=\arg\max_{k\neq y} f_k(\mathbf{r})$ the strongest competing class. With $\mathbf{w}=\boldsymbol{\beta}_y-\boldsymbol{\beta}_{k^\star}$ and $b=b_y-b_{k^\star}$, we define
\begin{equation}
\mathrm{BPI}(\mathbf{r})=\frac{\mathbf{w}^\top\mathbf{r}+b}{\left|\mathbf{w}^\top\mathbf{c}_{\hat k}+b\right|},
\label{eq:bpi}
\end{equation}
where $\mathbf{c}_k$ is the mean representation of the datapoints of class $k$ that the probe classifies correctly (the \emph{refined centroid}; see Appendix~\ref{app:setup}). BPI is the signed distance to the pairwise hyperplane separating the true class from its highest-logit rival, normalized by the absolute distance of the predicted-class refined centroid to that same hyperplane; it is positive exactly when the datapoint is classified correctly. Thus, $\mathrm{BPI}=-1$ means that a misclassified point has the same signed pairwise margin as this centroid. BPI measures position relative to a classifier-defined hyperplane; it does not measure proximity to a class cluster or local class density. The selected hyperplane need not be the nearest boundary of the multiclass classifier. This normalization provides a dimensionless reference for describing points across representations. Equal BPI values indicate equal margins relative to the respective centroid references, but do not necessarily indicate equal statistical difficulty, class density, or distance measured relative to within-class variability.

\paragraph{Rescue and breakage rates.}
We compare a network before and after training, using the same architecture and initialization seed and evaluating a linear probe on the test set at each layer in both cases. A datapoint misclassified before training but correctly classified afterward is \emph{rescued}; one that changes from correctly to incorrectly classified is \emph{broken}. For a group of datapoints, the \emph{rescue rate} is the fraction of initially misclassified points that are rescued, and the \emph{breakage rate} is the fraction of initially correctly classified points that are broken.

\begin{figure}[t]
\begin{center}
\safefig{0.9\linewidth}{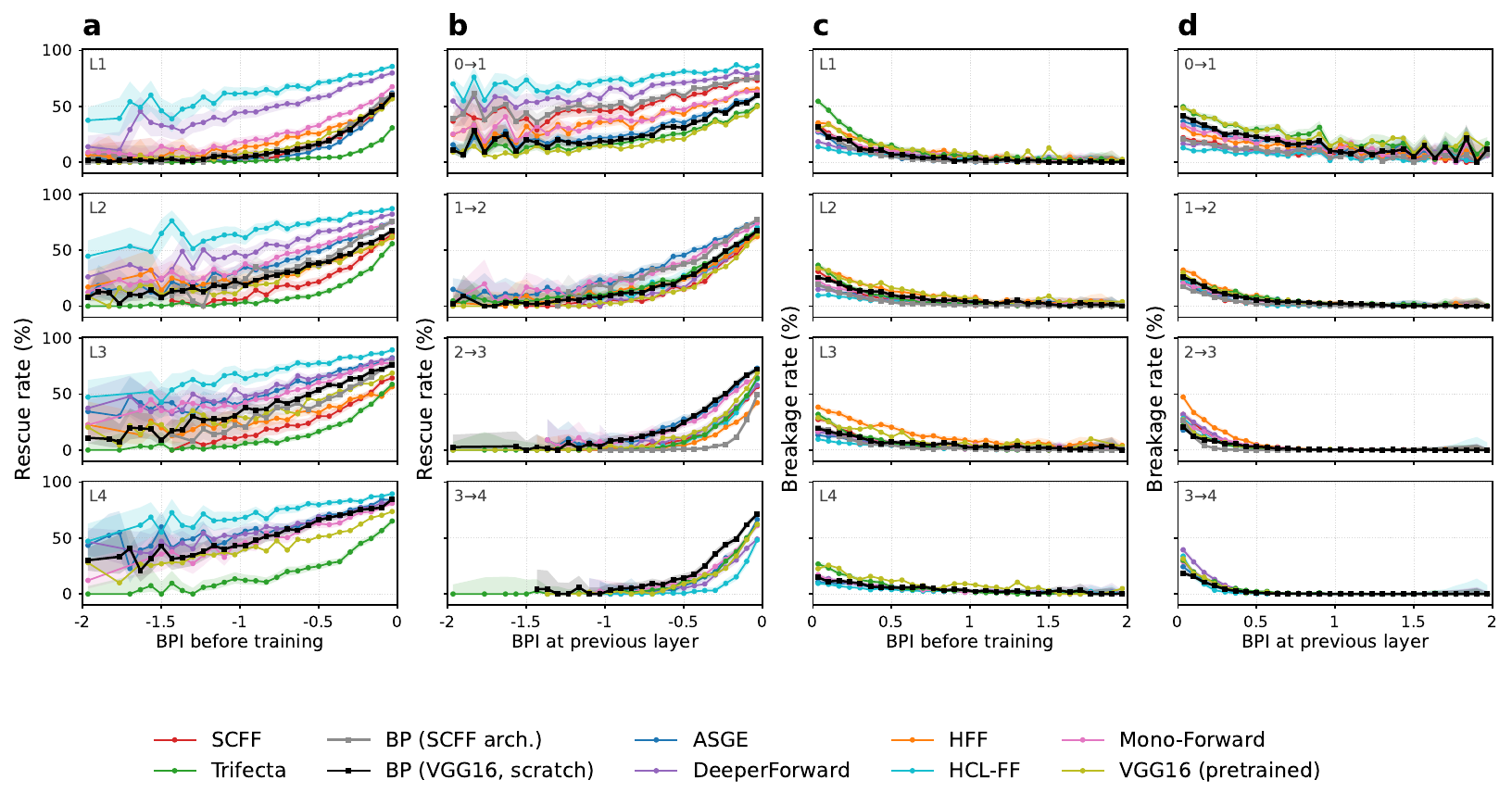}
\caption{Rescue and breakage rates against boundary proximity on CIFAR-10 (test set), one curve per method. Each point is the mean rescue rate (a, b) or breakage rate (c, d) over the datapoints of one BPI bin (30 equal-width bins on $[-2,0)$ for rescue and on $[0,2)$ for breakage, plotted at the bin center; the outermost bin also collects datapoints beyond $-2$ or $2$). Rows are the first four layers (a, c) or layer transitions (b, d); the first transition, $0{\to}1$, compares a linear probe on the input pixels with the first layer. Shaded bands are nominal 95\% Wilson intervals from pooled example-level counts, a descriptive binwise uncertainty that ignores dependencies across seeds and networks (Appendix~\ref{app:setup}); bins with fewer than 20 datapoints are omitted. \textbf{(a)}~Across training: fraction of datapoints misclassified by the untrained network's probe at layer $\ell$ that are correct after training, against BPI in the untrained network. \textbf{(b)}~Across depth: fraction misclassified at layer $\ell$ of the trained network that are correct at layer $\ell{+}1$, against BPI at layer $\ell$. \textbf{(c,d)}~Breakage for the same comparisons (BPI is positive for correctly classified points). BP: backpropagation on the SCFF architecture, or VGG16 trained from scratch (three seeds). For VGG16 (pretrained), ``training'' is ImageNet pretraining (one set of weights, so no band). Every layer is shown in \cref{fig:all_c10}, and the same figure for CIFAR-100, Tiny ImageNet and ImageNet-1K in Appendix~\ref{app:datasets}.}
\label{fig:rate}
\end{center}
\end{figure}

\Cref{fig:rate}a,c show rescue and breakage rates as functions of untrained BPI for each layer and method. We group datapoints by their BPI under the probe of the \emph{untrained} network, capturing their positions before learning. We use 30 equal-width bins (width $1/15$) on each of the intervals $[-2,0)$ for initially misclassified datapoints (negative BPI) and $[0,2)$ for initially correctly classified datapoints, with the outermost bins also collecting datapoints beyond these ranges. Each point on a curve in \Cref{fig:rate} shows the rescue or breakage rate within a BPI bin and is plotted at the bin center; bins containing fewer than 20 datapoints are omitted.

Across all methods, the probability that training rescues an initially misclassified datapoint or breaks an initially correctly classified datapoint generally decreases the farther the datapoint lies from the decision boundary before training, as measured by its untrained BPI (\Cref{fig:rate}a). This trend also holds on CIFAR-100 and Tiny ImageNet, as well as in the exploratory ImageNet-1K experiment (Appendix~\ref{app:datasets}). The dependence is particularly steep for methods that gain little from training beyond the untrained baseline: at layer 3, SCFF rescues about 7\% of datapoints with $\mathrm{BPI}<-1$, compared with about 50\% of those with $-0.5<\mathrm{BPI}<0$; the corresponding rates for HFF are 22\% and 48\%. Backpropagation on the same architecture as SCFF achieves higher rescue rates in both groups (15\% and 64\%) but exhibits the same dependence of rescue and breakage rates on untrained BPI. Methods that gain the most from training (HCL-FF, ASGE, Mono-Forward, and VGG16 trained from scratch or pretrained) shift the entire curve upward and rescue a substantial fraction of deeply embedded points at layer 3 (24--60\% with $\mathrm{BPI}<-1$), yet still rescue a larger fraction of points near the boundary (58--85\% with $-0.5<\mathrm{BPI}<0$). The dependence is sharpest at the first layer: SCFF, backpropagation, ASGE, and both VGG16 networks rescue fewer than 5\% of datapoints with $\mathrm{BPI}<-1$. HCL-FF is the exception, plausibly because its first analyzed unit is a block of several layers. Breakage is likewise concentrated near the decision boundary and decreases to nearly zero beyond $\mathrm{BPI}\approx 1$ for all the methods (\cref{fig:rate}c). Together, the rescue and breakage curves show that a datapoint's position before training is strongly associated with whether its classification changes: points initially far from the probe's boundary are less likely to be either rescued or broken.

\subsection{Across-Depth Rescue Depends on Boundary Proximity at the Preceding Depth}
\label{sec:across_depth}

Having examined how classifications change with training at a fixed layer, we now examine how they change across successive layers within a fully trained network. We fit a linear probe at every depth, compute each datapoint's BPI at layer $\ell$ using the corresponding probe, and examine whether a datapoint misclassified at layer $\ell$ becomes correctly classified at layer $\ell{+}1$ (rescued), or whether an initially correctly classified datapoint becomes misclassified (broken).

 Using the same BPI binning procedure, we group datapoints by their BPI at the preceding layer and compute the mean rescue and breakage indicators within each bin. \Cref{fig:rate}b shows the resulting rescue rates, with remarkably similar curves across methods and layer transitions: the rate remains low until the datapoint is within roughly $0.5$ to $1$ of the boundary, then rises steeply. Pooled over the layer transitions of each network on CIFAR-10, a datapoint with $\mathrm{BPI}<-1$ at the previous layer is rescued at most 11\% of the time for every method, whereas one within $0.5$ of the boundary is rescued about 33 to 50\% of the time. Breakage is confined to datapoints just on the correct side of the boundary (\cref{fig:rate}d): roughly 11--29\% of datapoints with $0<\mathrm{BPI}<0.25$ become misclassified after passing through an additional layer, compared with at most 0.6\% of those with $\mathrm{BPI}>1$. Thus, an additional layer generally rescues points with strongly negative BPI at the preceding depth less often than those nearer the boundary. This pattern also holds on CIFAR-100 and Tiny ImageNet (Appendix~\ref{app:datasets}). The first transition, from the input pixels to the first layer (row $0{\to}1$), is the exception: pixel-space BPI is a much weaker predictor of rescue and breakage, plausibly because the first layer substantially changes the representation. On CIFAR-10, probe accuracy rises from 41\% on the input pixels to 44--84\% at layer 1, depending on the method. Datapoints with pixel-space $\mathrm{BPI}<-1$ are rescued 11--68\% of the time, compared with 36--83\% for those with $-0.5<\mathrm{BPI}<0$. Meanwhile, 4--16\% of datapoints correctly classified by the pixel-space probe with $\mathrm{BPI}>1$ are broken.

\begin{figure}[t]
\begin{center}
\safefig{\linewidth}{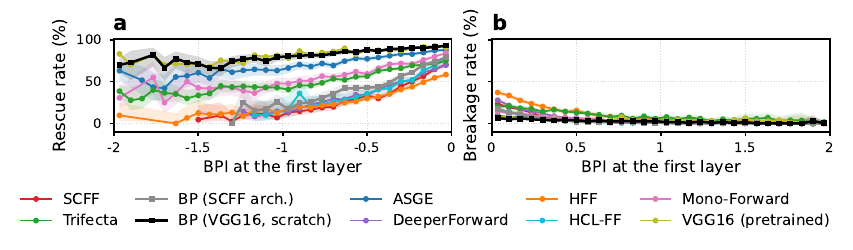}
\caption{Rescue (a) and breakage (b) from the first to the deepest layer of the trained network, on CIFAR-10 (test set), against BPI at the first layer. In (a) the datapoints are misclassified by the first-layer probe and the rate is the fraction correct at the method's deepest layer (depths in \cref{tab:setup}); in (b) they are correct at the first layer and the rate is the fraction misclassified at the deepest layer. Curves, bands and omitted bins as in \cref{fig:rate}. CIFAR-100 and Tiny ImageNet: \cref{fig:f2l_c100,fig:f2l_tin}.}
\label{fig:first_to_last}
\end{center}
\end{figure}

We also examine rescue and breakage from the first to the last layer to assess how strongly a datapoint's final classification depends on its BPI at the first layer. In \Cref{fig:first_to_last}, we group datapoints by their first-layer BPI and compute their mean rescue and breakage rates at the last layer. For every method, the rescue rate increases with first-layer BPI, although this dependence is weak for the two VGG16 networks. The methods separate into two groups. For SCFF, backpropagation on the SCFF architecture, DeeperForward, HCL-FF, and HFF, at most 18\% of datapoints with $\mathrm{BPI}<-1$ at the first layer are correctly classified at the last layer, compared with 46--64\% of those with $-0.5<\mathrm{BPI}<0$. For ASGE, Mono-Forward, Trifecta, and both VGG16 networks, a substantial fraction of deeply misclassified datapoints are eventually rescued (41--77\%), although their rescue rates remain lower than those of points near the boundary (68--90\%). Thus, although points far inside an incorrect decision region are generally less likely to be rescued by the next layer, the first-to-last comparison shows that substantial rescue can accumulate across depth in some methods despite low rescue rates at individual transitions.  On CIFAR-100 and Tiny ImageNet, we also observe similar behavior (Appendix~\ref{app:datasets}).

In the above experiments (\Cref{fig:first_to_last}, \cref{fig:rate}b,d), the linear probes are trained on concatenated representations from layers $1$ through $\ell$ at each depth $\ell$. The across-depth analysis therefore measures the additional predictive value of a layer given access to earlier features. It does not measure the quality of that layer's output in isolation, and concatenation can preserve access to useful earlier features when later representations deteriorate. Appendix~\ref{app:noncat} repeats the analysis using individual-layer probes without concatenating features from previous layers. The qualitative dependence of rescue on BPI persists in this setting, although some quantitative results change.

\subsection{Control Experiments to Separate the Effect of Learning from Probe Noise and Random Features}
\label{sec:null}

A dependence of rescue rate on BPI could arise even without representation learning: a linear probe refitted on slightly different features may flip its predictions mainly near the decision boundary. To establish what such non-learning sources produce, we repeat the rescue-rate-by-BPI analyses under three null conditions in which no representation is trained. These nulls give the rescue-vs-BPI and breakage-vs-BPI curves expected from random features, depth, and probe fitting alone; we attribute to learning only the amount by which a trained method's curves exceed these baselines, in particular for points deep inside incorrect decision regions.

\begin{figure}[t]
\begin{center}
\safefig{0.9\linewidth}{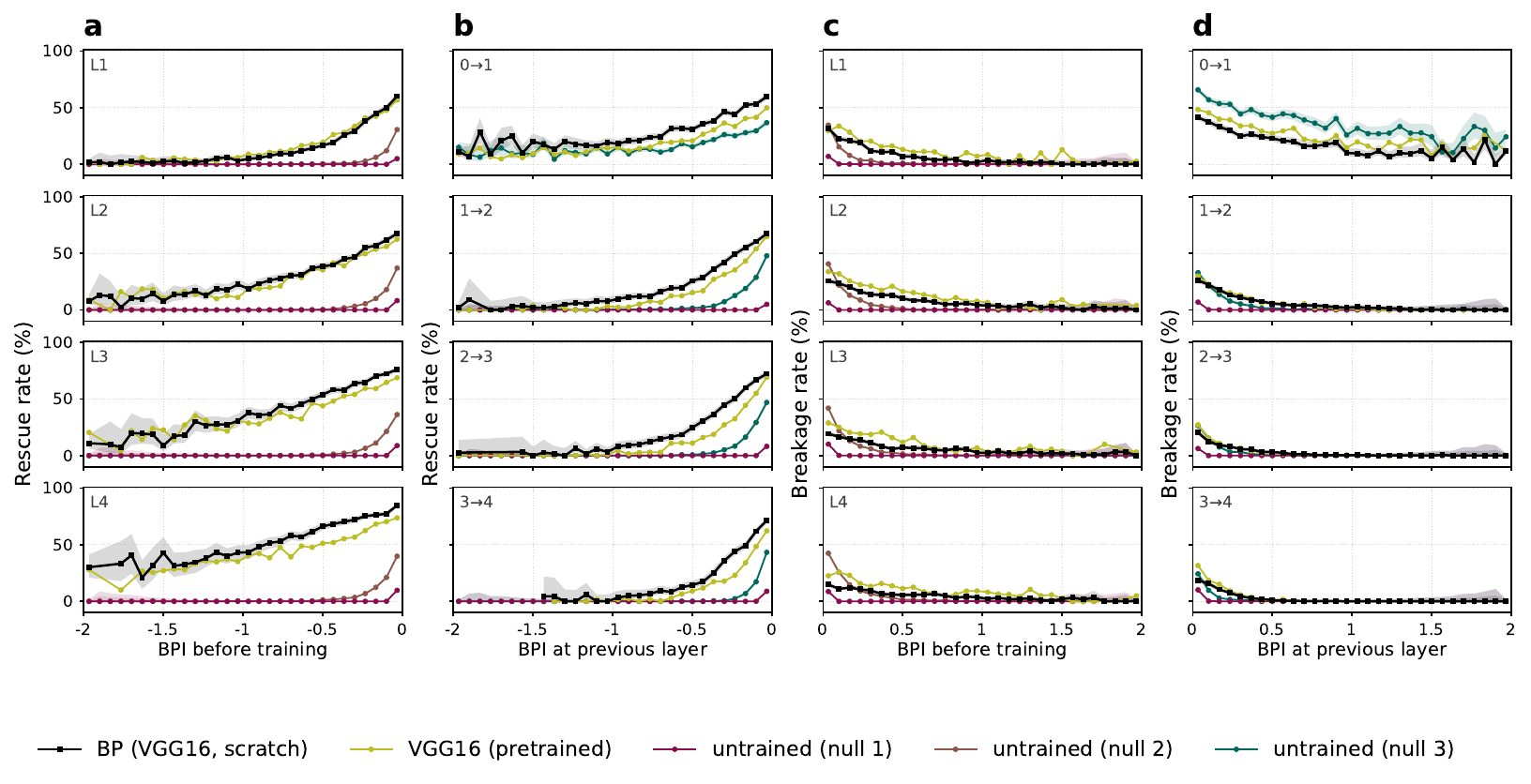}
\caption{Null controls for VGG16 on CIFAR-10, in the format of \cref{fig:rate} (first four layers; every layer is shown in \cref{fig:null_vgg16_all}). Trained: VGG16 from scratch on CIFAR-10 and pretrained VGG16; the other curves come from the untrained VGG16 (three initializations). Null~1: two probes with different seeds on the same untrained features (none for row $0{\to}1$). Null~2 (a,c): untrained network $A$ against $B$ (all ordered pairs). Null~3 (b,d): layers $\ell$ and $\ell{+}1$ of one untrained network (row $0{\to}1$: pixels and layer~1). The controls for every other method are in Appendix~\ref{app:null}, and those for ImageNet-1K in \cref{fig:all_in1k}.}
\label{fig:null_rate}
\end{center}
\end{figure}

\begin{figure}[t]
\begin{center}
\safefig{\linewidth}{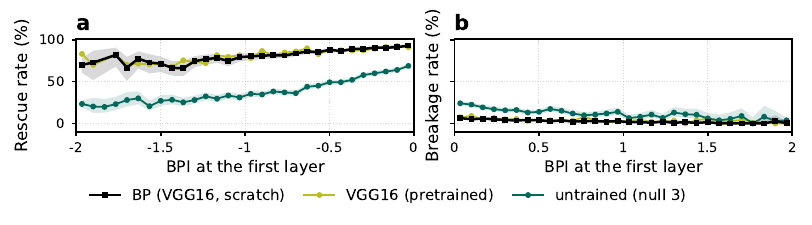}
\caption{Null controls for the first-to-last comparison of \cref{fig:first_to_last}, for VGG16 on CIFAR-10. The untrained curve uses the first and the deepest layer of the same untrained network (null~3). Colors as in \cref{fig:null_rate}. Other methods: Appendix~\ref{app:null}.}
\label{fig:null_f2l}
\end{center}
\end{figure}


\emph{Null~1 (probe noise).} Two linear probes with different seeds are fitted on the same untrained features; a datapoint misclassified by the first is ``rescued'' if the second classifies it correctly, and datapoints are binned by their BPI under the first probe. The probes are nearly identical: they differ on only 0.4 to 3.1\% of the test datapoints, all next to the boundary, and their BPIs correlate at 0.989 to 0.998 (Appendix~\ref{app:null3}). Probe noise therefore rescues almost no datapoint, except for a small rate (below about 10\%) in the bins next to the boundary, and breaks few correct ones (Null~1 curves in \cref{fig:null_rate}, other methods and datasets in Appendices~\ref{app:null} and~\ref{app:null_ds}), therefore probe noise accounts for little of the rescue and breakage.

\emph{Null~2 (across training).} The trained network is replaced by a second untrained network with a different seed; a datapoint misclassified on the first network is rescued if the probe on the second classifies it correctly, binned by BPI in the first network. A second random initialization rescues essentially no datapoint farther than about $0.5$ from the boundary, and rises only near it, well below the trained networks (\cref{fig:null_rate}a, Appendices~\ref{app:null} and~\ref{app:null_ds}). The trained networks instead rescue a sizeable fraction of datapoints at all BPI levels, including those deep inside incorrect decision regions. The rise toward the boundary is thus generic to a refitted probe and far exceeds probe noise, but only training lifts the rescue rate across the whole BPI range. Breakage near the boundary is also generic: a second initialization breaks about as many correct datapoints there as training does (\cref{fig:null_rate}c). The other methods, datasets (Appendices~\ref{app:null} and~\ref{app:null_ds})) show the same.

\emph{Null~3 (across depth).} Successive layers of one untrained network are compared: a datapoint misclassified at layer $\ell$ is rescued if it is correct at layer $\ell{+}1$, binned by BPI at layer $\ell$. Here the untrained curves lie close to the trained ones: like them, they rescue almost nothing far from the boundary and rise steeply only near it, and in the first transition (from pixels) the untrained layer rescues about as many datapoints as the trained ones (\cref{fig:null_rate}b). Breakage is likewise similar, and in the first transition an untrained layer breaks more correct datapoints than a trained one (\cref{fig:null_rate}d). That a single layer rarely rescues far points (\cref{sec:across_depth}) is therefore not specific to training. The gain from training accumulates over depth (\cref{fig:null_f2l}): from the first to the deepest layer, the trained networks rescue most of the datapoints misclassified at the first layer, almost independently of their first-layer BPI, whereas an untrained network rescues far fewer of them, especially far from the boundary, and breaks more of the correct ones. Methods that gain least from training (HFF and SCFF, \cref{sec:untrained}) stay close to the untrained network, while the others exceed it clearly (Appendices~\ref{app:null} and~\ref{app:null_ds}).

\subsection{Backpropagation Exploits Depth Better Than Most Forward-Forward Methods, Whose Performance Saturates Within a Few Layers}
\label{sec:depth_profile}

\begin{figure}[t]
\begin{center}
\safefig{\linewidth}{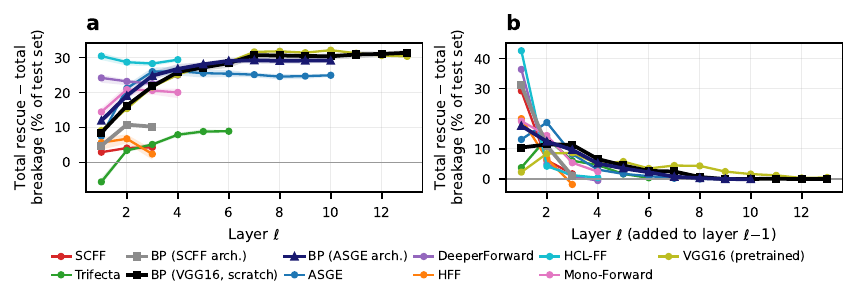}
\caption{Where in depth training builds linear separability, on CIFAR-10 (test set; mean over seeds, bands are the minimum and maximum over seeds). Total rescue minus total breakage, in \% of the test set, which equals the change in probe accuracy. \textbf{(a)}~Across training, at layer $\ell$: the probe accuracy of the trained network minus that of the untrained network of the same seed. \textbf{(b)}~Across depth, at layer $\ell$: the probe accuracy of the trained network at depth $\ell$ minus that at depth $\ell{-}1$ (for $\ell=1$, minus the raw-pixel probe), i.e.\ the layer added to the earlier ones. BP (ASGE arch.): backpropagation on ASGE's ten-block architecture. Other curves as in \cref{fig:rate}. CIFAR-100 and Tiny ImageNet: \cref{fig:net_rescue_c100,fig:net_rescue_tin}.}
\label{fig:net_rescue}
\end{center}
\end{figure}

Summed over all datapoints, total rescue minus total breakage is the change in probe accuracy, and plotting it against depth shows where in the network linear separability is built. We use this metric in \cref{fig:net_rescue} to compare how much accuracy training adds at each depth, and thus how well each method, local or backpropagation-based, exploits depth (other datasets: Appendix~\ref{app:datasets}). Across training (\cref{fig:net_rescue}a), the methods differ mainly in how early they reach their final gain. The shallow local methods (DeeperForward, HCL-FF, Mono-Forward and SCFF) obtain most of their gain at the first layer and gain little beyond the third (\cref{fig:net_rescue}b). Backpropagation on SCFF's three-layer architecture gains more than SCFF at every layer and builds its gain more gradually. ASGE, with ten layers, and the VGG16 networks also obtain only a minor part of their final gain at the first layer, and Trifecta is even below its untrained network there. The cleanest comparison is backpropagation on ASGE's own architecture. Its gain matches that of ASGE through the first four layers, but ASGE then plateaus, whereas backpropagation keeps gaining over several more layers and ends clearly higher: in this comparison, backpropagation continues to benefit from additional depth after the gains from local training have largely saturated. The picture is the same on CIFAR-100 and Tiny ImageNet (\cref{fig:net_rescue_c100,fig:net_rescue_tin}), where backpropagation on ASGE's architecture again ends well above ASGE.

\cref{fig:net_rescue} (other datasets: Appendix~\ref{app:datasets}) shows that several local methods achieve larger first-layer gains than the backpropagation-trained networks, whereas backpropagation distributes its gains more gradually across depth. This pattern is consistent with local objectives emphasizing immediate improvements at each layer, while end-to-end optimization allows improvements to emerge at later depths. Comparisons with deeper BP-trained networks therefore remain informative when a shallower local method has already plateaued over its later evaluated layers: an important advantage of BP in these experiments is its ability to exploit additional depth, rather than necessarily achieving larger gains in shallow networks.

\section{Conclusion}
\label{sec:conclusion}

Final accuracy alone does not reveal how much training adds beyond an untrained architecture or which datapoints benefit. In the evaluated networks, untrained convolutional features provide substantial linear-probe accuracy, while layer-wise ablations show considerable variation in the benefit of training individual layers. BPI-conditioned rescue curves show that initially misclassified points nearer the probe's boundary generally have higher rescue probabilities. This dependence on boundary proximity also occurs without learning; the magnitude of rescue and its accumulation across depth distinguish the evaluated methods, and backpropagation exploits depth better than most local methods, whose gains saturate within a few layers. Some local methods rescue a substantial fraction of points far inside incorrect decision regions, leaving these regions as a useful evaluation target rather than an established inability of local learning. Our findings are descriptive and conditional on the evaluated architectures, datasets (CIFAR-10, CIFAR-100, Tiny ImageNet, and an exploratory ImageNet-1K analysis), training configurations, and probe protocol. They do not establish the mechanisms responsible for the observed rescue patterns.

\bibliography{references}
\bibliographystyle{style_1}

\appendix
\renewcommand{\thefigure}{A.\arabic{figure}}
\setcounter{figure}{0}
\renewcommand{\thetable}{A.\arabic{table}}
\setcounter{table}{0}

\pagefig{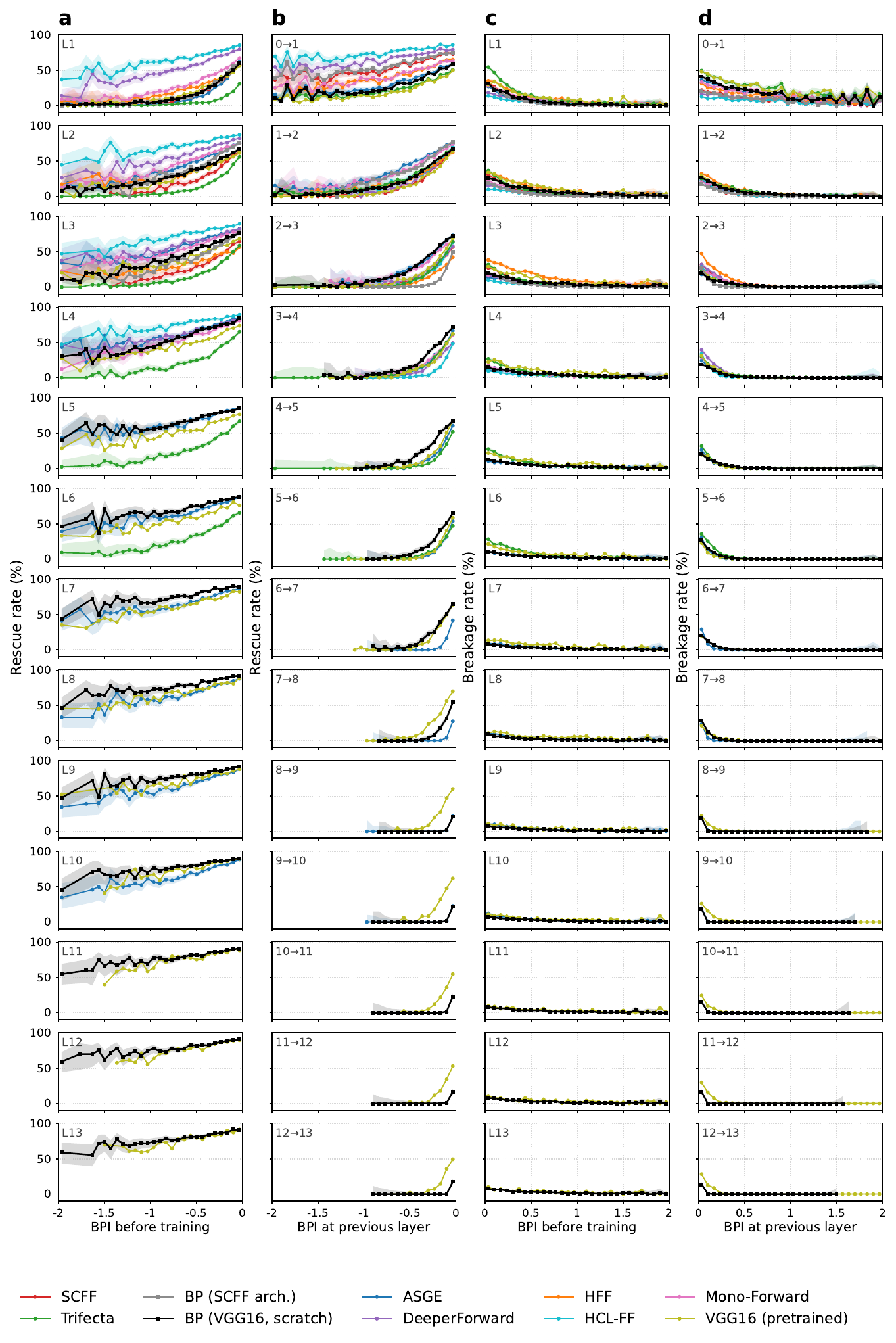}{As \cref{fig:rate}, with a row for every layer.}{fig:all_c10}
\pagefig{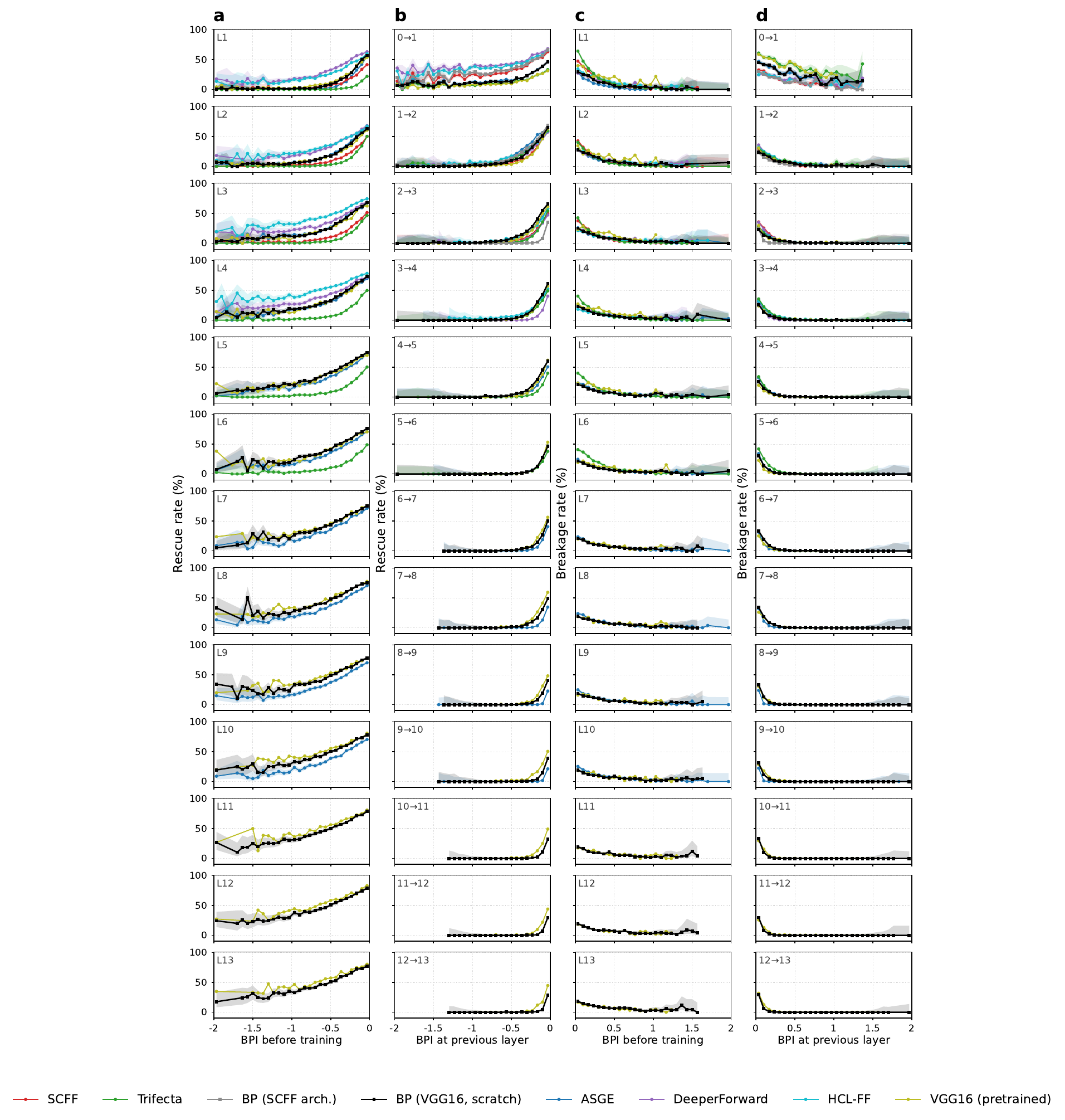}{As \cref{fig:rate}, on CIFAR-100, with a row for every layer.}{fig:all_c100}
\pagefig{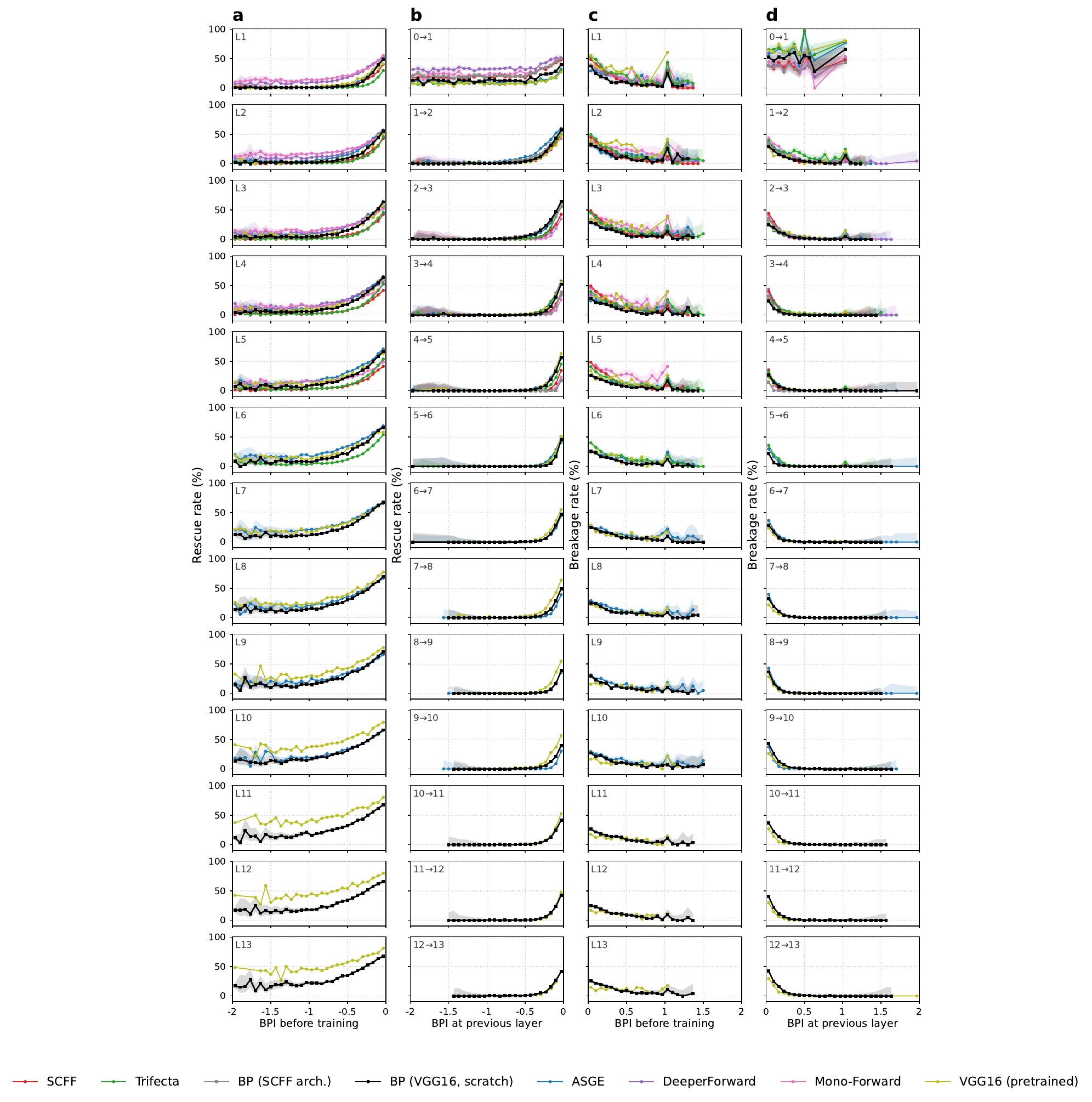}{As \cref{fig:rate}, on Tiny ImageNet, with a row for every layer.}{fig:all_tin}
\pagefig{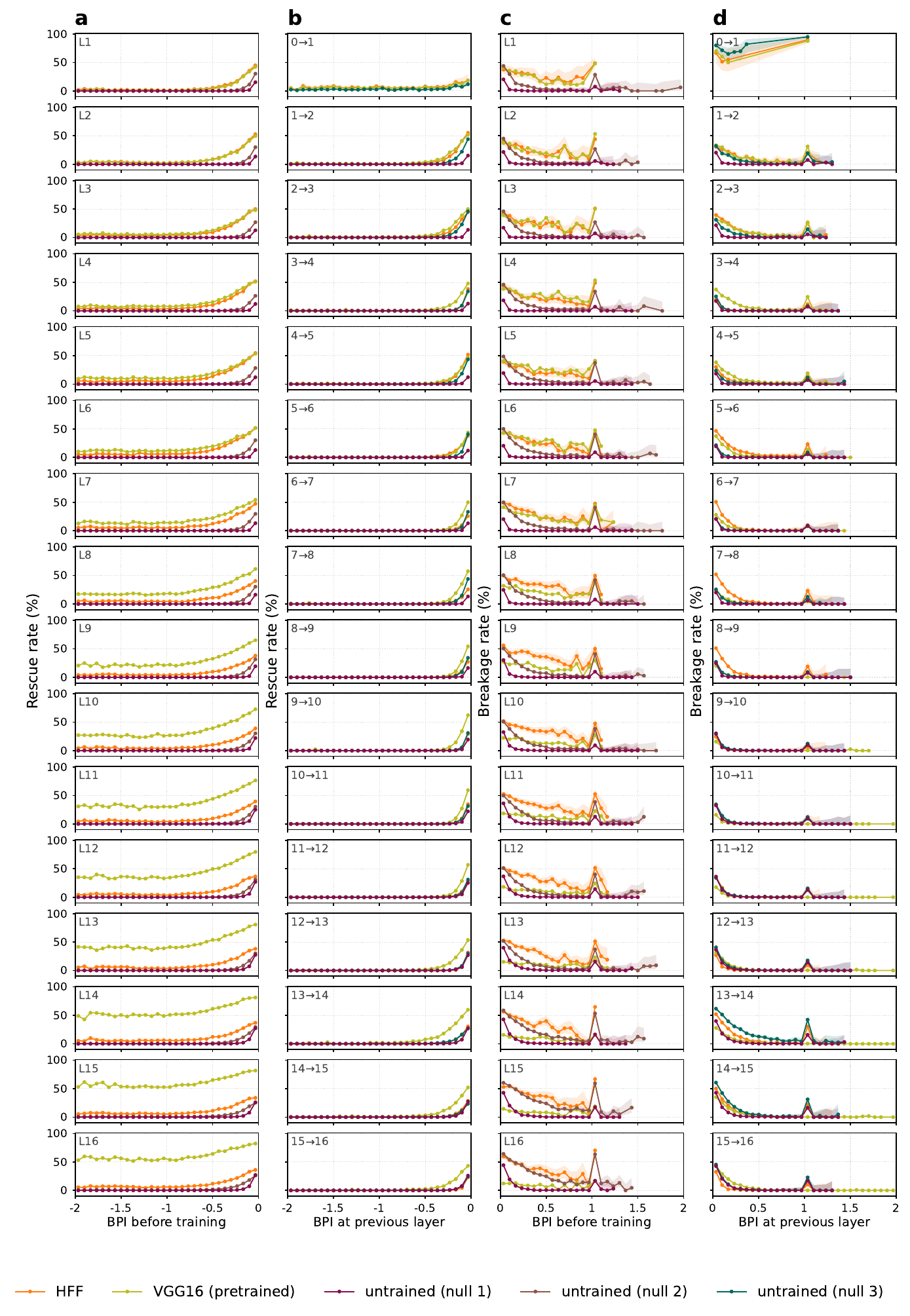}{As \cref{fig:rate}, on ImageNet-1K, with a row for every layer: HFF-VGG16 (our training), pretrained VGG16 and untrained VGG16 (nulls 1 to 3 of \cref{sec:null}; Appendix~\ref{app:datasets}).}{fig:all_in1k}

\begin{figure}[t]
\begin{center}
\safefig{\linewidth}{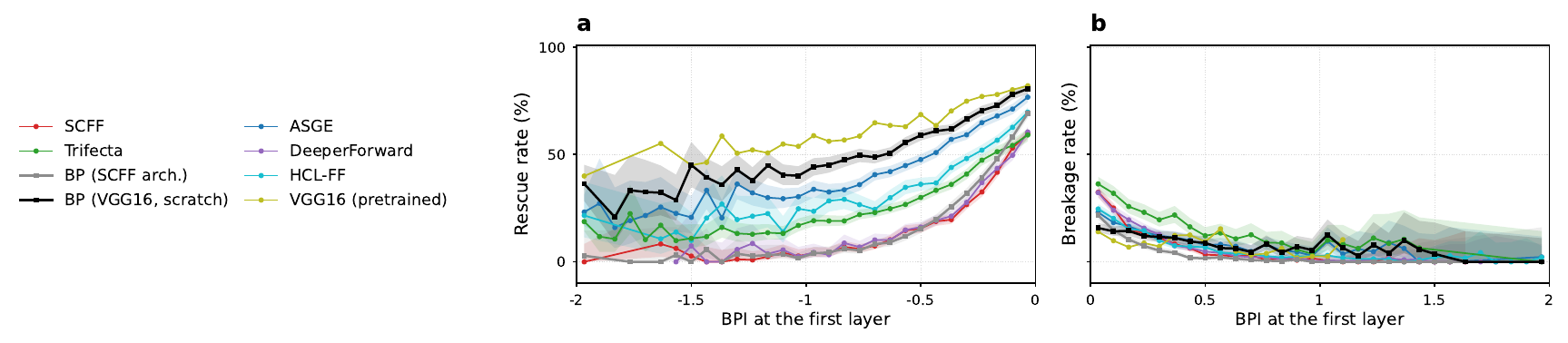}
\caption{As \cref{fig:first_to_last}, on CIFAR-100.}
\label{fig:f2l_c100}
\end{center}
\end{figure}

\begin{figure}[t]
\begin{center}
\safefig{\linewidth}{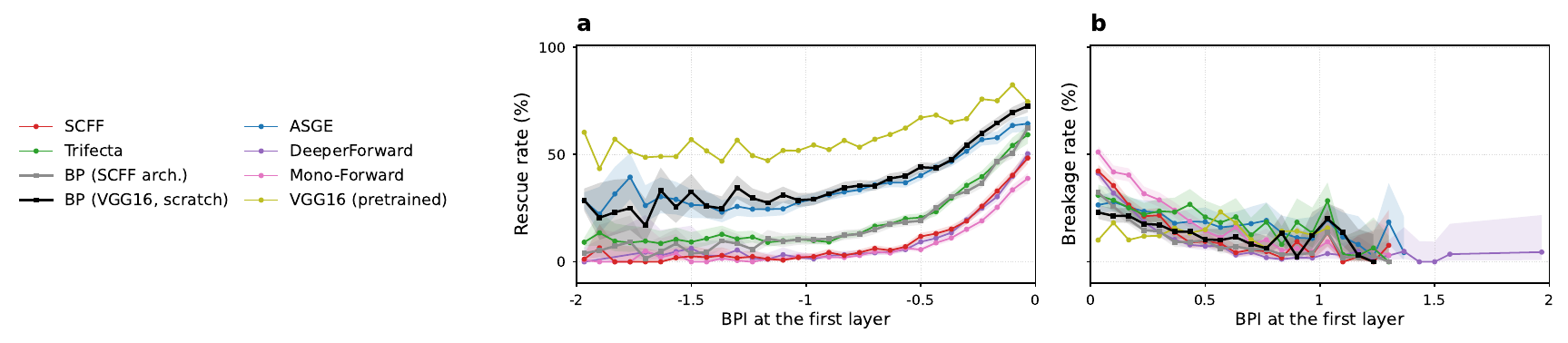}
\caption{As \cref{fig:first_to_last}, on Tiny ImageNet.}
\label{fig:f2l_tin}
\end{center}
\end{figure}

\begin{figure}[t]
\begin{center}
\safefig{\linewidth}{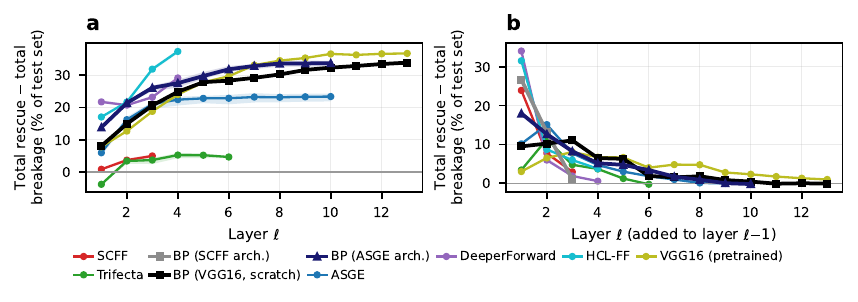}
\caption{As \cref{fig:net_rescue}, on CIFAR-100 (BP on the SCFF architecture is available only across depth).}
\label{fig:net_rescue_c100}
\end{center}
\end{figure}

\begin{figure}[t]
\begin{center}
\safefig{\linewidth}{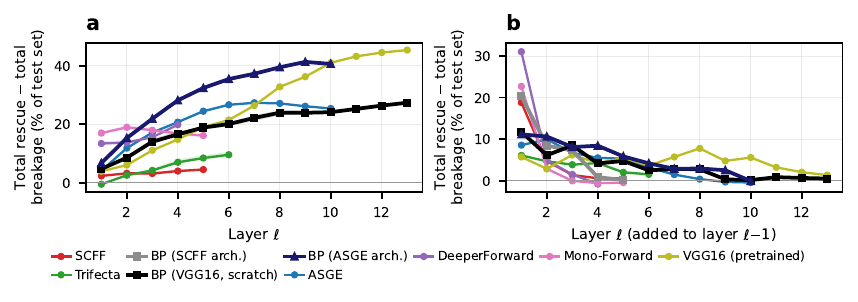}
\caption{As \cref{fig:net_rescue}, on Tiny ImageNet (BP on the SCFF architecture is available only across depth).}
\label{fig:net_rescue_tin}
\end{center}
\end{figure}

\section{Experimental Setup}
\label{app:setup}

\paragraph{Methods and datasets.}
\Cref{tab:setup} lists the methods. All experiments use the standard training and test splits of CIFAR-10 and CIFAR-100~\citep{krizhevsky2009learning}; for Tiny ImageNet~\citep{le2015tiny} the official validation set is used as the test set. Each network is trained on the training set with up to three seeds, except the pretrained VGG16, which is a single set of weights. Not every method was run on every dataset; methods that were not run are absent from the corresponding figure. Methods with public code are run from that code; the others are re-implemented from their papers (\cref{tab:setup}). BP (VGG16, scratch) is the torchvision VGG16 convolutional stack without batch norm on $32\times32$ inputs ($64\times64$ on Tiny ImageNet, with a linear read-out on the $2\times2\times512$ output), trained on each dataset with SGD (cosine schedule, 100 epochs, random crops and flips) to about 91.7\% (CIFAR-10), 64.9\% (CIFAR-100) and 44.8\% (Tiny ImageNet) test accuracy. BP (ASGE arch.) is the ten-block ASGE architecture trained end to end with a single cross-entropy loss on a linear read-out of the last block, with the same recipe (93.1\% test accuracy on CIFAR-10, 71.4\% on CIFAR-100 and 59.5\% on Tiny ImageNet); it starts from the same initial weights as ASGE, so its untrained network is ASGE's. The pretrained VGG16 reads inputs upsampled to $224\times224$.

\begin{table}[!ht]
\caption{Methods analyzed. ``Layers'' is the number of probed layers on CIFAR-10 (SCFF, BP and Mono-Forward use five layers on Tiny ImageNet); for DeeperForward and HCL-FF a layer is one residual block. BP is backpropagation on the SCFF architecture with a linear classification layer; the pretrained VGG16 is compared with its random initialization and the from-scratch VGG16 with the initialization it was trained from.}
\label{tab:setup}
\centering\footnotesize
\begin{tabular}{p{3.5cm}p{3.3cm}cp{2.1cm}}
\toprule
Method & Local objective & Layers & Implementation \\
\midrule
SCFF~\citep{chen2025self} & self-contrastive goodness & 3 & official code \\
Trifecta~\citep{dooms2024the} & symmetric goodness loss, batch norm (no overlapping updates) & 6 & re-implemented \\
DeeperForward~\citep{sun2025deeperforward} & layer-normalized mean goodness & 4 & official code \\
HCL-FF~\citep{yao2026hclff} & hierarchical goodness + supervised contrastive & 4 & official code \\
HFF~\citep{sarode2026hyperspherical} & class prototypes on the hypersphere & 3 & re-implemented \\
Mono-Forward~\citep{gong2025monoforward} & layer-wise cross-entropy, detached & 4 & re-implemented \\
ASGE~\citep{gong2025asge} & spatial goodness, VGG11-style & 10 & official code \\
BP (SCFF arch.) & end-to-end cross-entropy & 3 & --- \\
BP (VGG16, scratch)~\citep{simonyan2015very} & end-to-end cross-entropy & 13 & own training \\
BP (ASGE arch.) & end-to-end cross-entropy & 10 & own training \\
VGG16 (pretrained)~\citep{simonyan2015very} & ImageNet-pretrained (BP) & 13 & pretrained weights \\
\bottomrule
\end{tabular}
\end{table}

\paragraph{Representations and probes.}
For every method the representation at depth $\ell$ is the concatenation of the pooled activations of layers $1,\dots,\ell$, so that representations are nested and a probe at depth $\ell{+}1$ can reproduce the one at depth $\ell$. This is the same protocol for all methods and does not follow each method's own read-out. Each layer's activation is taken as the method's feature extractor provides it and pooled to a small grid: SCFF as in its own read-out (its configured max or average pooling), and adaptive average pooling to a $2{\times}2$ grid for the other methods ($4{\times}4$ for HFF on CIFAR-10; $1{\times}1$ for ASGE on Tiny ImageNet); each pooled map is then standardized per sample. Linear softmax probes are fitted on frozen training-set representations, separately for each depth, for the trained and the untrained network. Every probe is a dropout layer (rate 0.1) followed by a linear layer, trained with cross-entropy and Adam (learning rate $10^{-3}$, weight decay $10^{-4}$, batch size 2048) on precomputed features, without augmentation, with the learning rate halved at 20, 35, 50, 60, 70, 80 and 90\% of the epochs and the random seed set to 42 before each fit; BPI uses the weights of the linear layer. The number of epochs is 200 by default (across-depth probes, and the across-training probes of SCFF on CIFAR-10 and CIFAR-100, VGG16 and backpropagation), 150 for the across-training probes of ASGE, Trifecta, DeeperForward, HCL-FF, Mono-Forward, HFF and SCFF on Tiny ImageNet, and 100 on ImageNet-1K. The probe for the row $0{\to}1$ of \cref{fig:rate} is a linear softmax probe on the flattened, mean/std-normalized input pixels, fitted with the same protocol as the other probes; its BPI is the same for every method. The untrained network is the same architecture with the same seed and random initialization; for the ablation of \cref{fig:ablation}, layers beyond the first $m$ keep this initialization. In \cref{fig:ablation}, ASGE, Trifecta, DeeperForward, HFF, HCL-FF and Mono-Forward use the architectures of \cref{tab:setup}, whereas the SCFF networks are the published architectures extended by two convolutional layers (five layers for CIFAR-10 and CIFAR-100, seven for Tiny ImageNet). For the six methods other than SCFF, we construct each ablation by retaining the first $m$ layers from the fully trained network and resetting subsequent layers to their original initialization. This is equivalent to training only the retained layers when their updates and training states are independent of downstream layers.

\paragraph{BPI.}
BPI is computed on test datapoints from the probe fitted on the training set. The refined centroid $\mathbf{c}_k$ in \cref{eq:bpi} is the mean of the test representations of class $k$ that the probe classifies correctly, following the convention of the original analysis. The refined centroids are estimated from correctly classified test examples, so the normalization depends on the probe and the evaluation sample. This use of evaluation labels makes BPI a retrospective diagnostic rather than a predictor available at test time; it does not affect the fitted probe or its reported accuracy. The normalization may be sensitive to classes with few correctly classified examples or small centroid margins. Robustness to alternative centroid estimates and normalizations is not established here.

\paragraph{Rescue and breakage curves.}
For rescue, datapoints misclassified before are binned by BPI into 30 equal-width bins on $[-2,0)$, the first bin being open to $-\infty$; for breakage, datapoints correctly classified before are binned into 30 bins on $[0,2)$, the last bin being open to $+\infty$. Counts are pooled over seeds. Shaded bands are nominal 95\% Wilson intervals calculated from pooled example-level counts. Because the same test images recur across seeds, and the null comparisons reuse networks, these intervals do not account for all dependencies or training-run variability. They should be interpreted as descriptive binwise uncertainty rather than full confidence intervals over experimental randomness. Bins with fewer than 20 datapoints are omitted, and no band is drawn for the pretrained VGG16 because it is a single set of weights. The rows of \cref{fig:rate} show the first four layers (transitions) of each network; \cref{fig:all_c10,fig:all_c100,fig:all_tin,fig:all_in1k} show every layer.

\paragraph{Untrained networks for the null controls.}
For Null~1 the paper's probe (seed 42) is compared with a second probe fitted on the identical features with seed 43 (another initialization, mini-batch order and dropout masks); refitting the paper's probe reproduces the stored one exactly for SCFF and VGG16 and on 97 to 99\% of test datapoints for the other methods (GPU non-determinism), and the second seed changes the correctness of 1 to 3\% of the test datapoints at the deepest layer. For Null~2 the untrained networks of the different seeds of a method are paired (all six ordered pairs); for Null~3 each untrained network is analyzed on its own, with the same probes, BPI and binning as the trained networks. The untrained VGG16 is that of the pretrained pipeline (CIFAR-10 upsampled to $224\times224$, ImageNet normalization, bf16; seeds 1, 2 and 1234). Untrained networks are regenerated from the seed where the stored arrays do not cover them (ASGE seed~1 and the extra VGG16 seeds); they reproduce the per-depth accuracy of the layer-ablation baseline exactly (Trifecta, DeeperForward), and agree with stored untrained arrays on the correctness of 97 to 99\% of test datapoints (HFF, ASGE; BPI correlation $\geq 0.99$).

\section{Null Controls for Every Method}
\label{app:null}

\Cref{fig:null_vgg16_all,fig:null_scff,fig:null_bp_scffarch,fig:null_trifecta,fig:null_asge,fig:null_deeperforward,fig:null_hff,fig:null_hclff,fig:null_monoforward} repeat the null controls of \cref{sec:null} with every layer, for VGG16 and for every other method on CIFAR-10 (three seeds each); \cref{tab:null} summarizes them. BP (SCFF arch.) and SCFF share the same untrained networks, so their null curves coincide. \Cref{app:null_ds} repeats the controls on CIFAR-100 and Tiny ImageNet.

\pagefig{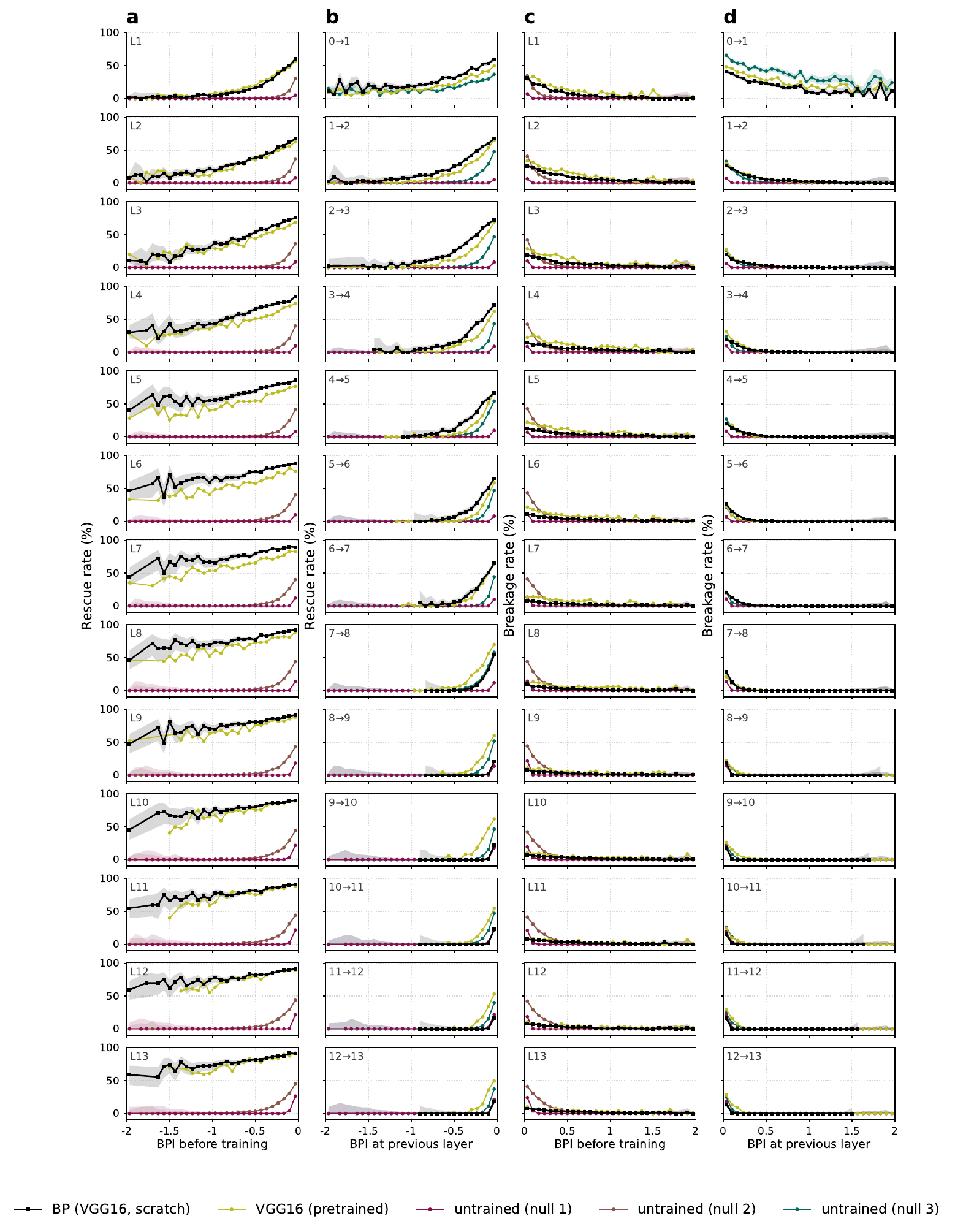}{Null controls for VGG16 on CIFAR-10: as \cref{fig:null_rate} with every layer.}{fig:null_vgg16_all}

\begin{table}[t]
\caption{Rescue rate (\%) on CIFAR-10 with seeds pooled, as ``trained / null 2 / null 1'' for the across-training column and ``trained / null 3 / null 1'' for the across-depth column. \emph{Across training, layer~3}: datapoints misclassified by the layer-3 probe of the untrained network, against BPI in that network. \emph{Across depth, $1{\to}2$}: datapoints misclassified at layer~1, rescued at layer~2 (null~1: layer~1 with a second probe). \emph{First$\to$last} (trained / null~3): all datapoints misclassified at the first layer that are correct at the deepest layer. Columns ``$\mathrm{BPI}<-1$'' and ``within 0.5'' restrict to datapoints with BPI below $-1$ or in about $[-0.5,0)$; they are computed on the 30 BPI bins of width $1/15$ used for the curves (bins with midpoint above $-0.5$), which differs from exact thresholds by less than 1 point.}
\label{tab:null}
\centering\scriptsize
\begin{tabular}{lccccc}
\toprule
 & \multicolumn{2}{c}{Across training, layer 3} & \multicolumn{2}{c}{Across depth, $1{\to}2$} & First$\to$last \\
\cmidrule(lr){2-3}\cmidrule(lr){4-5}\cmidrule(lr){6-6}
Method & $\mathrm{BPI}<-1$ & within 0.5 & $\mathrm{BPI}<-1$ & within 0.5 & all \\
\midrule
\input{null_table_rows}\\
\bottomrule
\end{tabular}
\end{table}

\nullpage{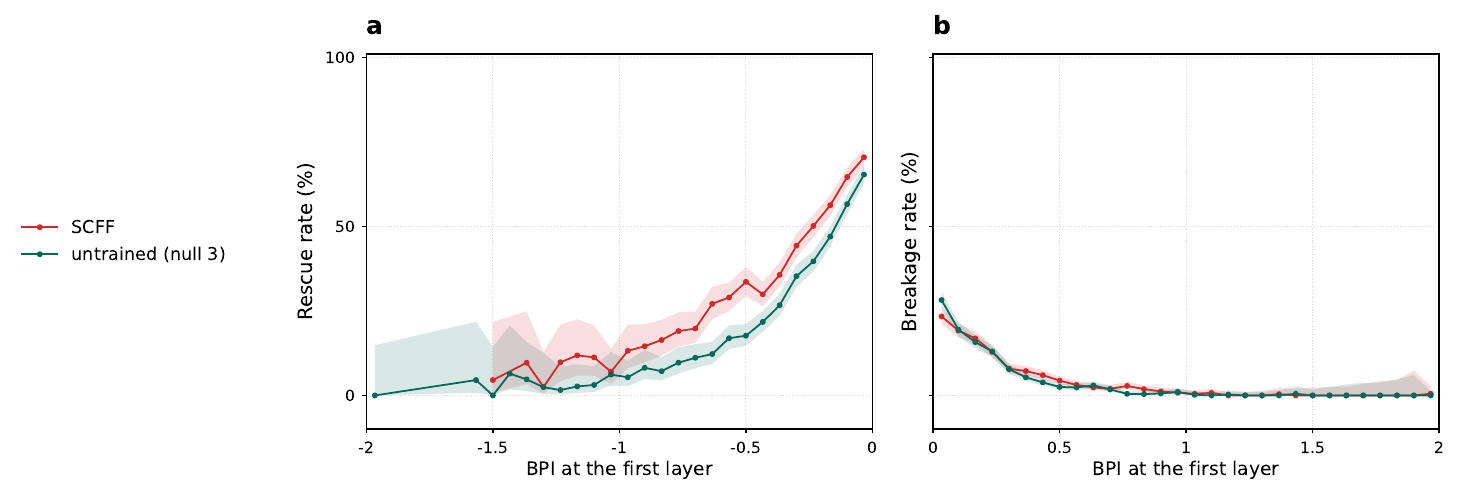}{SCFF}
\nullpage{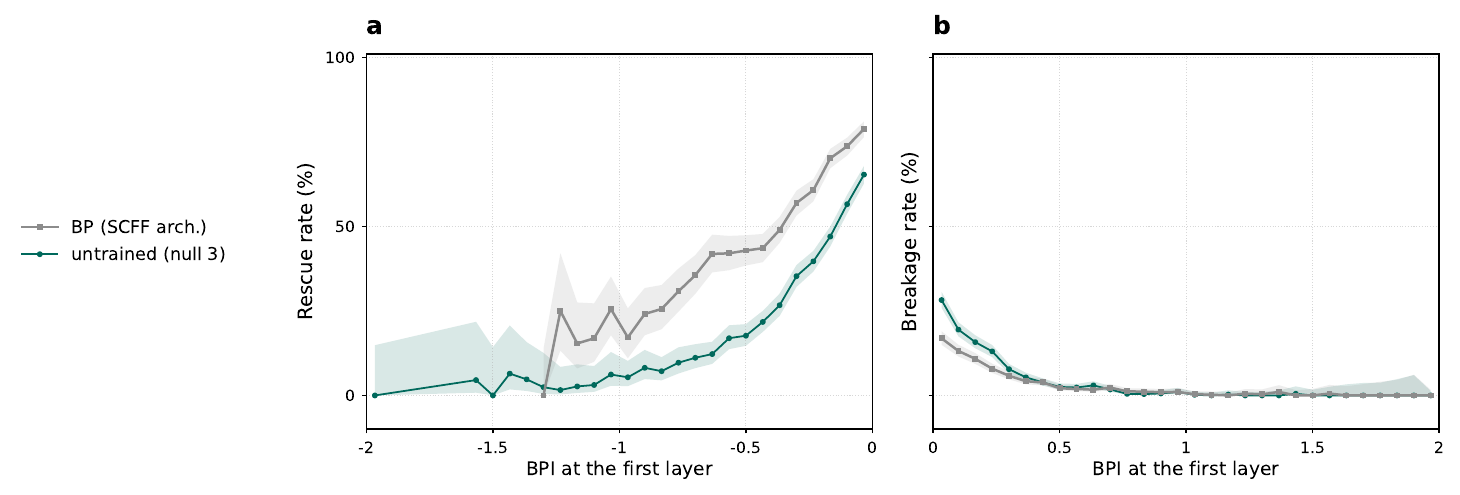}{BP (SCFF arch.)}
\nullpage{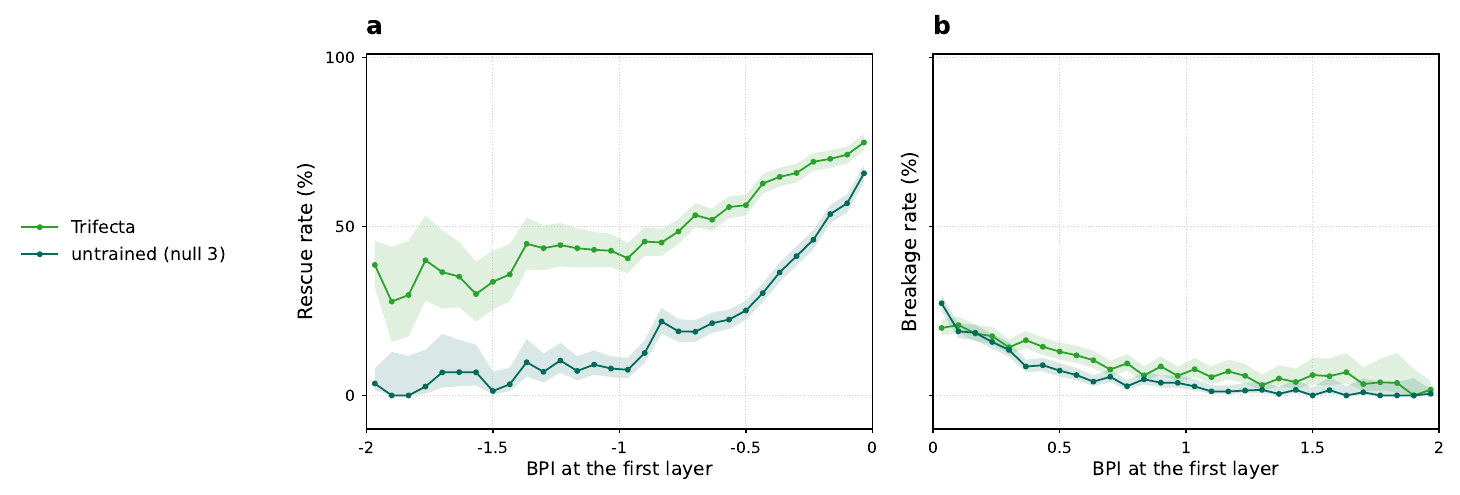}{Trifecta}
\nullpage{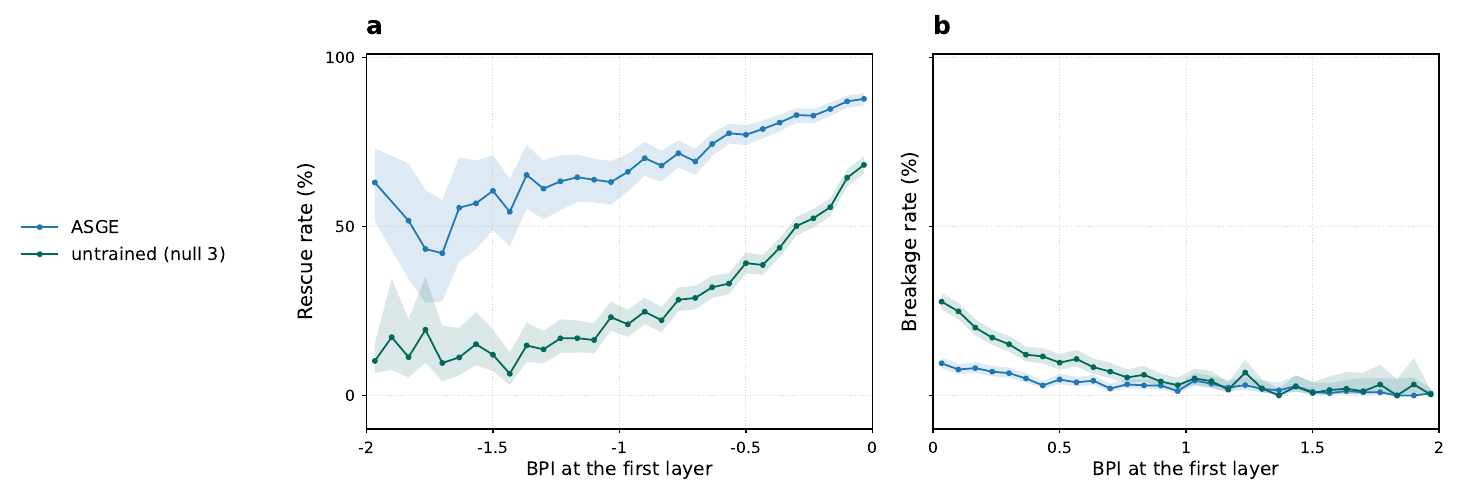}{ASGE}
\nullpage{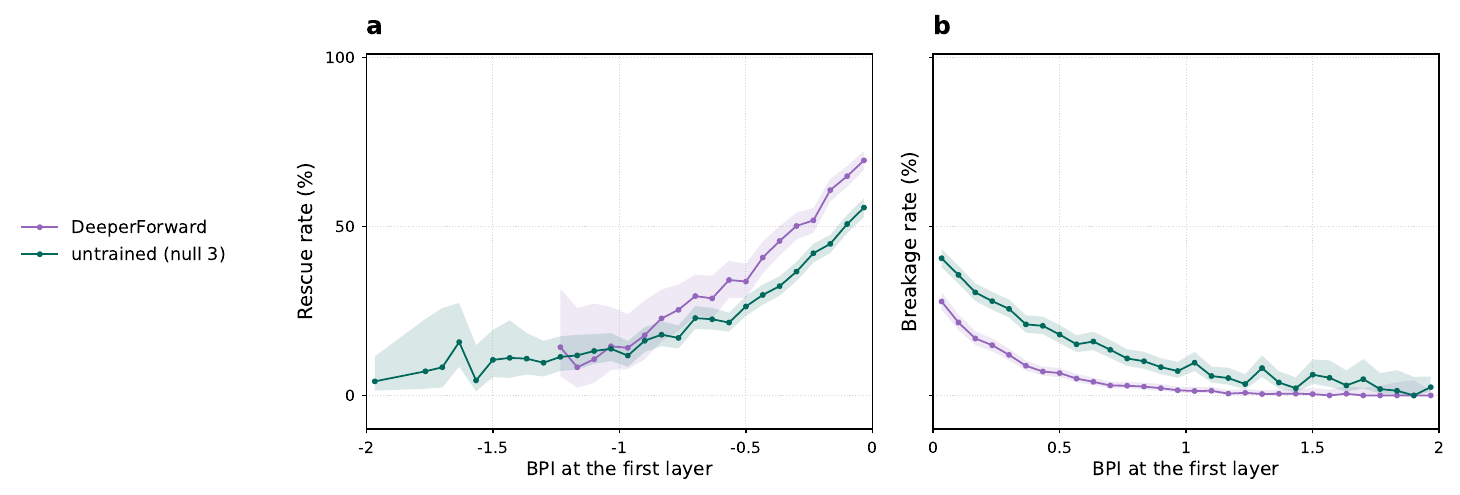}{DeeperForward}
\nullpage{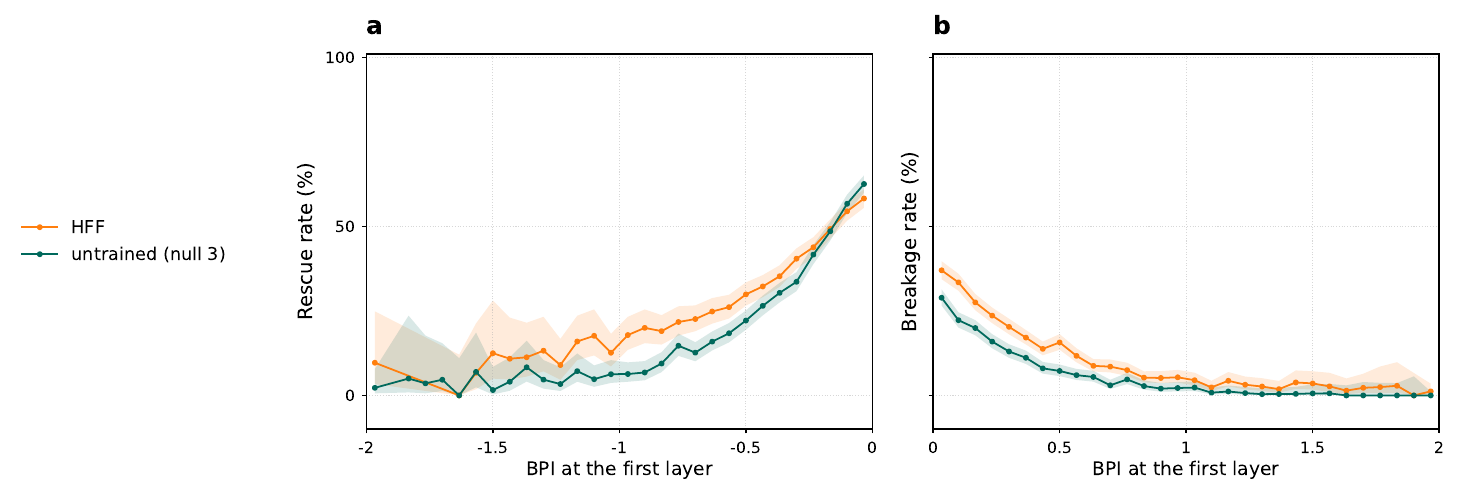}{HFF}
\nullpage{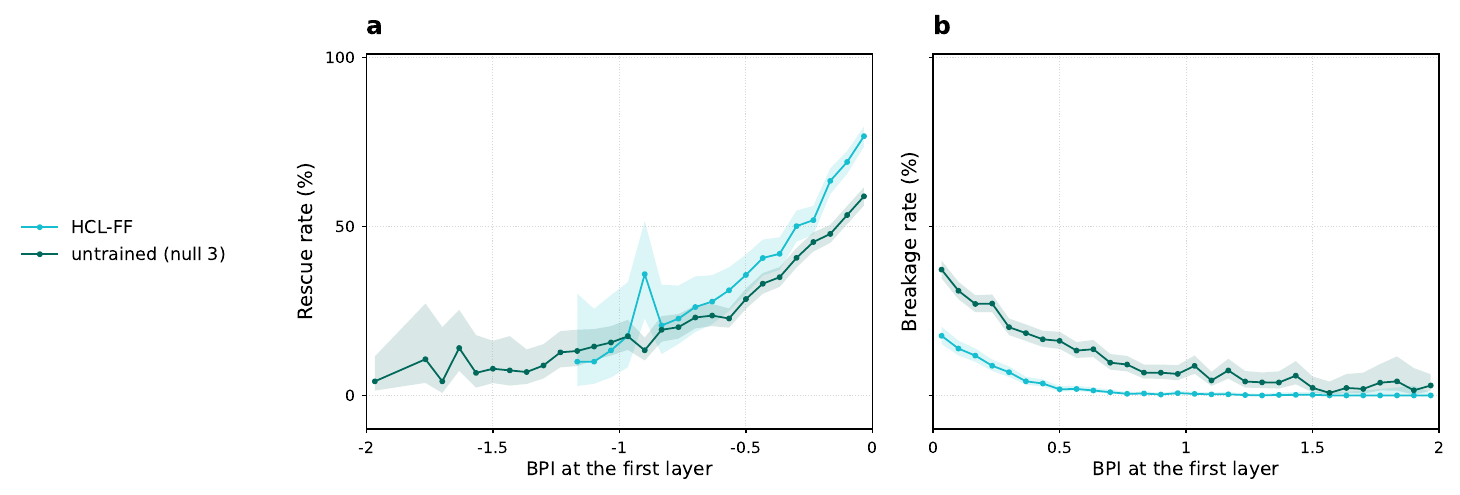}{HCL-FF}
\nullpage{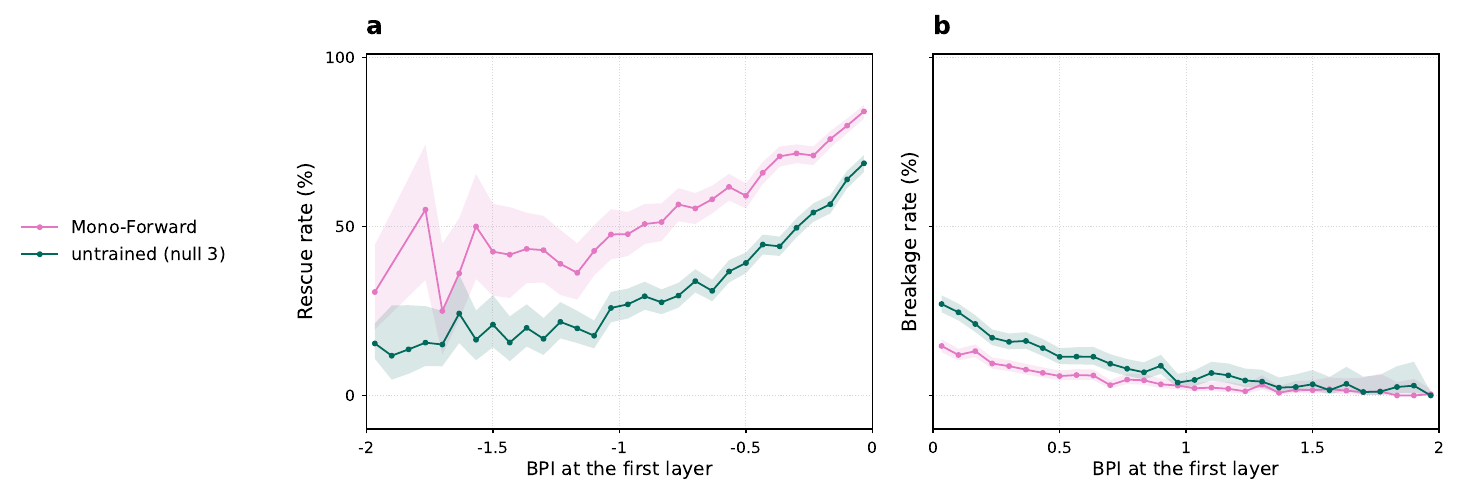}{Mono-Forward}

\section{Null Controls on CIFAR-100 and Tiny ImageNet}
\label{app:null_ds}

\Cref{sec:null} and Appendix~\ref{app:null} report the null controls on CIFAR-10; here they are repeated on CIFAR-100 and Tiny ImageNet, with the same three nulls, probes and binning. For VGG16 (\cref{fig:null_vgg16_cifar100,fig:null_vgg16_tinyimagenet}) the trained networks are VGG16 trained from scratch on the dataset (three seeds) and the ImageNet-pretrained VGG16, and the nulls are measured on the untrained VGG16 of the pretrained pipeline (three initializations, images upsampled to $224\times224$). For every other method run on the dataset, one page follows: the trained networks are its own (three seeds) and the untrained networks are its untrained networks of the same seeds. A method appears only on the datasets on which it was run (Appendix~\ref{app:setup}), and BP on the SCFF architecture, whose untrained networks are SCFF's, has no page of its own. Null~1 fits a second probe (seed 43) on the same untrained features, with the number of epochs of the paper's probe. The untrained networks are rebuilt from their seeds exactly as in each method's analysis, and refitting the paper's probe on them reproduces the stored predictions on 100\% of the test datapoints at the deepest depth in every network checked (34 checks); the two probes then disagree on 0.5 to 2.9\% of the test datapoints of the method networks at the deepest depth.

The controls behave as on CIFAR-10. For VGG16 at layer~3 across training, a second untrained network rescues 0.1\% of the datapoints with $\mathrm{BPI}<-1$ on both datasets, against 10.4 and 7.8\% (CIFAR-100) and 4.2 and 6.3\% (Tiny ImageNet) for the two trained networks, and 6.0 and 5.4\% of those within $0.5$ of the boundary, against 44 to 49\% and 31 to 35\%; probe noise rescues none of the former and at most 0.8\% of the latter. Across depth, at the transition $1{\to}2$ an untrained layer rescues 0.1\% of the datapoints with $\mathrm{BPI}<-1$ (trained: 0.1 to 0.6\%) and 10.2 and 8.3\% of those within $0.5$, against 24 to 32\% and 15 to 22\% for the trained networks. From the first to the deepest layer, the trained VGG16 networks rescue 59 and 66\% (CIFAR-100) and 44 and 60\% (Tiny ImageNet) of the datapoints misclassified at the first layer, against 27 and 13\% for the untrained network, which breaks 18 and 25\% of the correct ones, against 9 to 11\% and 13 to 17\% after training. The other methods behave alike (\cref{tab:null_ds}): at layer~3, no untrained network rescues more than 5\% of the datapoints with $\mathrm{BPI}<-1$ and probe noise rescues none, against 0.4 to 30\% (CIFAR-100) and 0.9 to 14\% (Tiny ImageNet) for the trained networks; within $0.5$ of the boundary a second untrained network rescues 6 to 23\% (CIFAR-100) and 5 to 17\% (Tiny ImageNet), against 22 to 63\% and 19 to 39\% after training, and probe noise at most 3.7\%. In the first-to-last comparison every trained network rescues more of the datapoints misclassified at the first layer than the untrained network of the same architecture, by 4 (SCFF, Tiny ImageNet) to 47 points (pretrained VGG16, Tiny ImageNet). At the transition $1{\to}2$, several trained layers rescue datapoints with $\mathrm{BPI}<-1$ hardly more often than an untrained layer (for example SCFF and DeeperForward on Tiny ImageNet), as on CIFAR-10.

\begin{table}[t]
\caption{Rescue rate (\%) of the null controls on CIFAR-100 and Tiny ImageNet, in the format of \cref{tab:null}: ``trained / null 2 / null 1'' for the across-training column and ``trained / null 3 / null 1'' for the across-depth column, seeds pooled; \emph{First$\to$last}: trained / null~3. For BP (VGG16, scratch) and VGG16 (pretrained) the nulls are those of the untrained VGG16 of the pretrained pipeline.}
\label{tab:null_ds}
\centering\scriptsize
\begin{tabular}{lccccc}
\toprule
 & \multicolumn{2}{c}{Across training, layer 3} & \multicolumn{2}{c}{Across depth, $1{\to}2$} & First$\to$last \\
\cmidrule(lr){2-3}\cmidrule(lr){4-5}\cmidrule(lr){6-6}
Method & $\mathrm{BPI}<-1$ & within 0.5 & $\mathrm{BPI}<-1$ & within 0.5 & all \\
\midrule
\multicolumn{6}{l}{\emph{CIFAR-100}}\\
\input{null_table_rows_cifar100}\\
\midrule
\multicolumn{6}{l}{\emph{Tiny ImageNet}}\\
\input{null_table_rows_tinyimagenet}\\
\bottomrule
\end{tabular}
\end{table}

\nullpageds{vgg16}{cifar100}{VGG16}{CIFAR-100}
\nullpageds{scff}{cifar100}{SCFF}{CIFAR-100}
\nullpageds{trifecta}{cifar100}{Trifecta}{CIFAR-100}
\nullpageds{asge}{cifar100}{ASGE}{CIFAR-100}
\nullpageds{deeperforward}{cifar100}{DeeperForward}{CIFAR-100}
\nullpageds{hclff}{cifar100}{HCL-FF}{CIFAR-100}
\nullpageds{vgg16}{tinyimagenet}{VGG16}{Tiny ImageNet}
\nullpageds{scff}{tinyimagenet}{SCFF}{Tiny ImageNet}
\nullpageds{trifecta}{tinyimagenet}{Trifecta}{Tiny ImageNet}
\nullpageds{asge}{tinyimagenet}{ASGE}{Tiny ImageNet}
\nullpageds{deeperforward}{tinyimagenet}{DeeperForward}{Tiny ImageNet}
\nullpageds{monoforward}{tinyimagenet}{Mono-Forward}{Tiny ImageNet}

\section{Agreement of the Two Probes in Null~1}
\label{app:null3}

\begin{figure}[!htbp]
\begin{center}
\safefig{\linewidth}{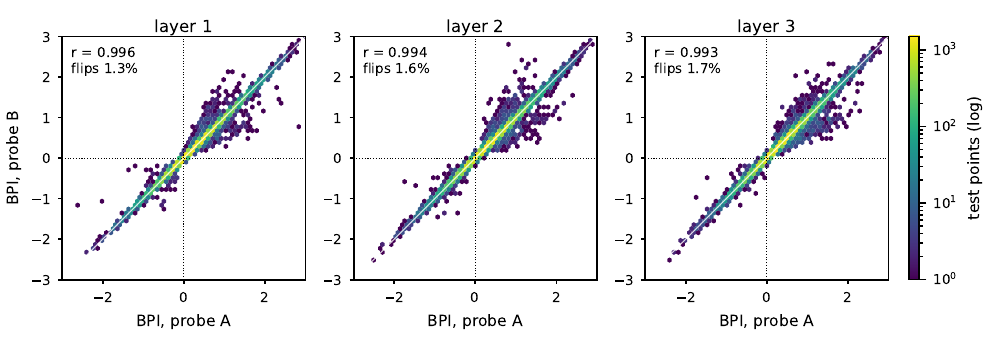}
\caption{Null~1, agreement of the two probes for SCFF (BP on the SCFF architecture shares these untrained networks): BPI under the paper's probe (A, seed 42) against BPI under a second probe (B, seed 43) fitted on the same untrained features, one panel per layer, test datapoints of the three untrained networks pooled. Log-count hexbins; white line: $y=x$; dotted lines: $\mathrm{BPI}=0$. $r$: Pearson correlation of the two BPIs; flips: share of test datapoints that the two probes classify differently.}
\label{fig:n3_scff}
\end{center}
\end{figure}

\begin{figure}[!htbp]
\begin{center}
\safefig{\linewidth}{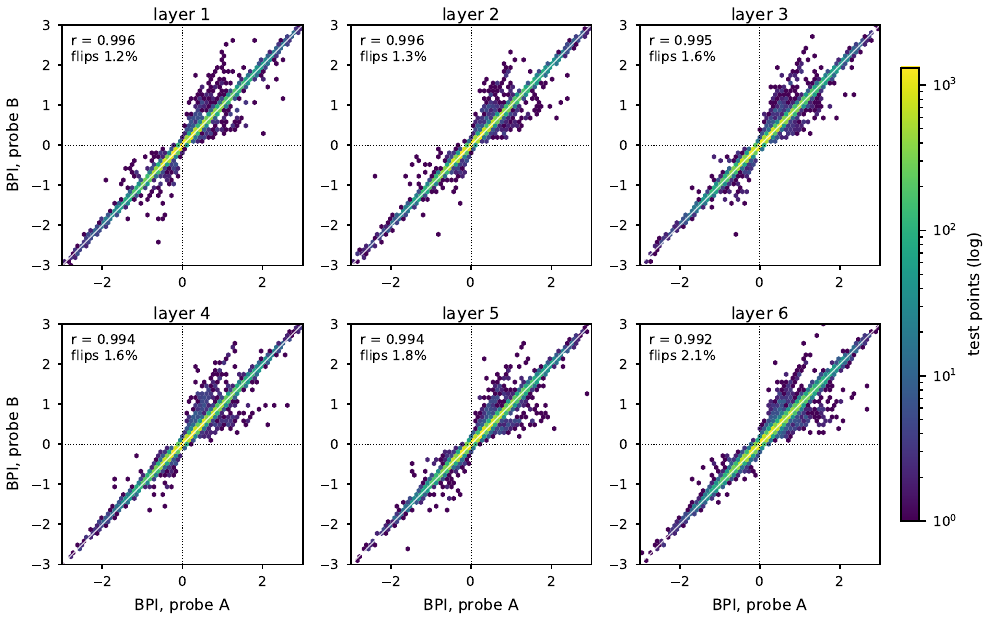}
\caption{As \cref{fig:n3_scff}, for Trifecta.}
\label{fig:n3_trifecta}
\end{center}
\end{figure}

\begin{figure}[!htbp]
\begin{center}
\safefig{\linewidth}{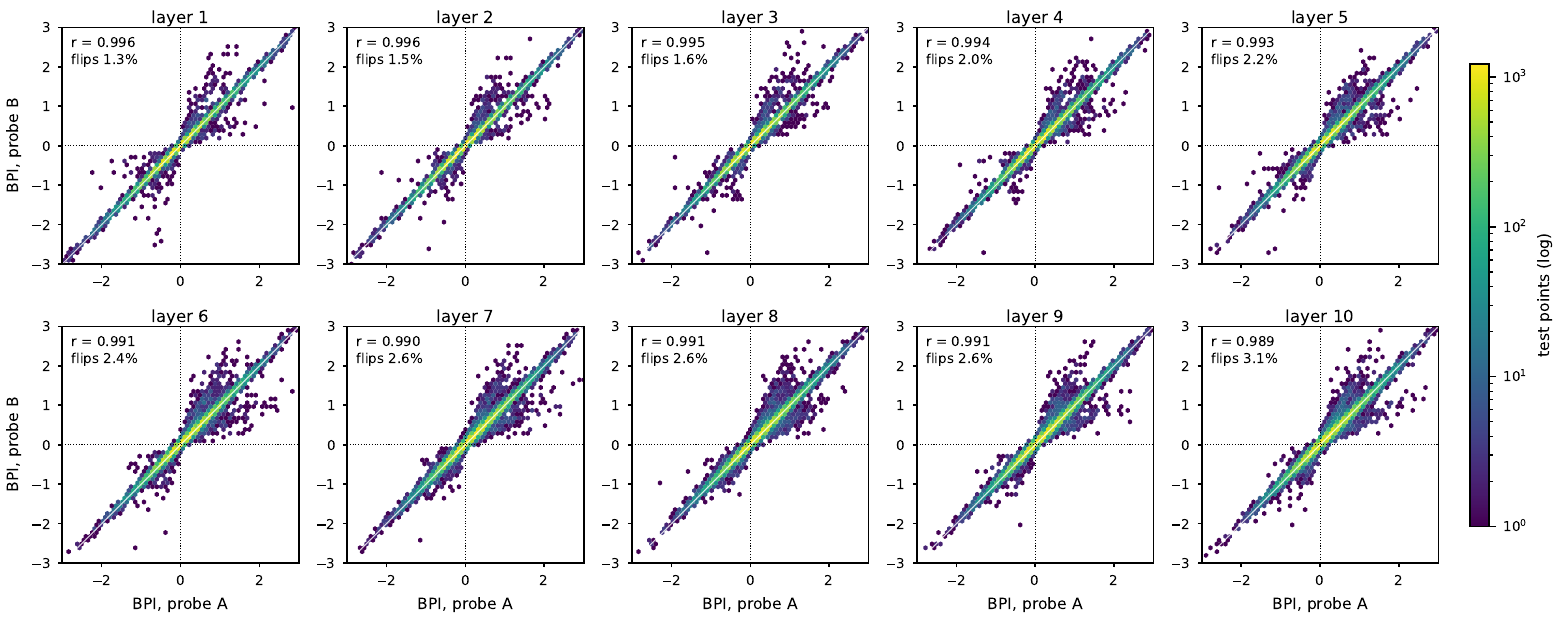}
\caption{As \cref{fig:n3_scff}, for ASGE.}
\label{fig:n3_asge}
\end{center}
\end{figure}

\begin{figure}[!htbp]
\begin{center}
\safefig{\linewidth}{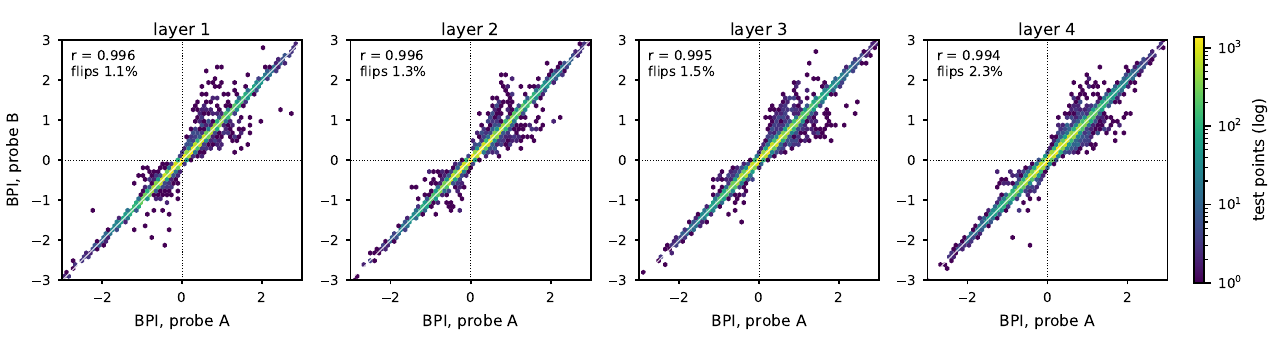}
\caption{As \cref{fig:n3_scff}, for DeeperForward.}
\label{fig:n3_deeperforward}
\end{center}
\end{figure}

\begin{figure}[!htbp]
\begin{center}
\safefig{\linewidth}{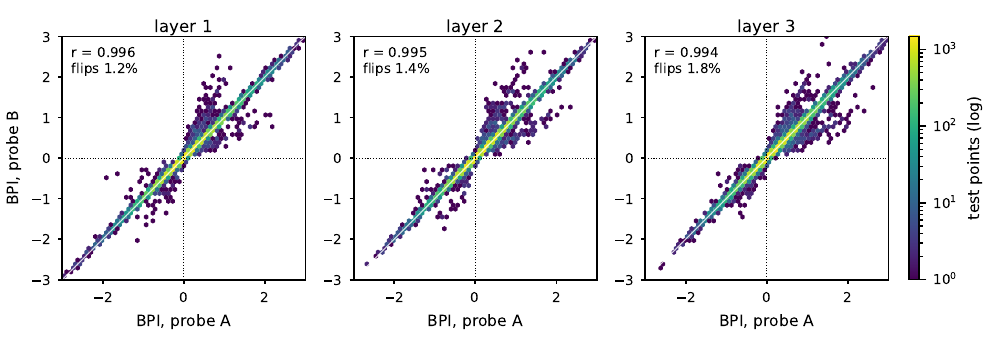}
\caption{As \cref{fig:n3_scff}, for HFF.}
\label{fig:n3_hff}
\end{center}
\end{figure}

\begin{figure}[!htbp]
\begin{center}
\safefig{\linewidth}{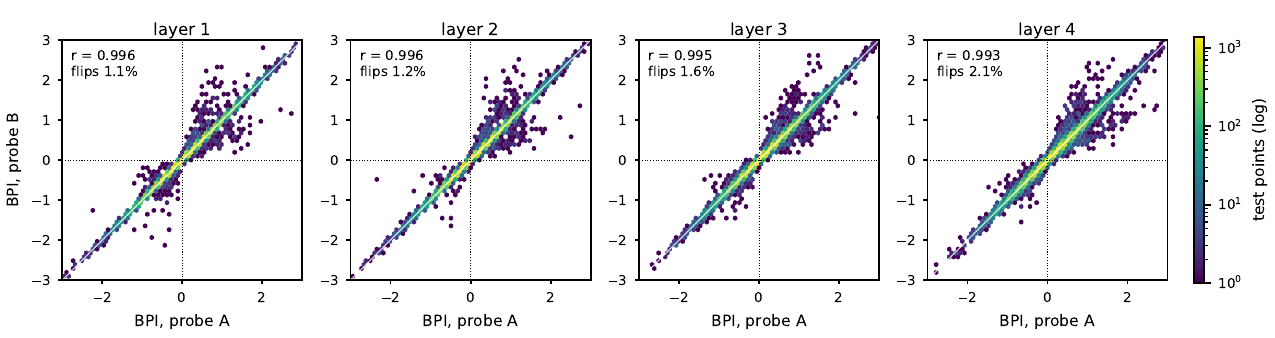}
\caption{As \cref{fig:n3_scff}, for HCL-FF.}
\label{fig:n3_hclff}
\end{center}
\end{figure}

\begin{figure}[!htbp]
\begin{center}
\safefig{\linewidth}{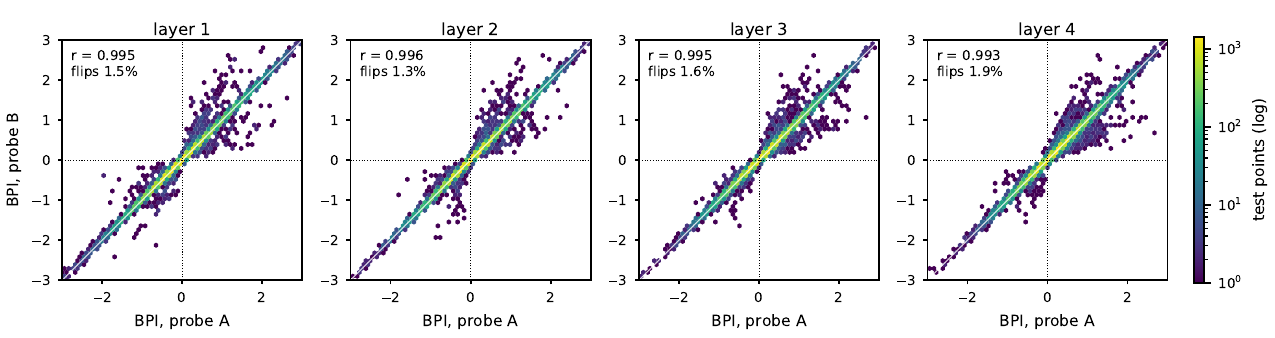}
\caption{As \cref{fig:n3_scff}, for Mono-Forward.}
\label{fig:n3_monoforward}
\end{center}
\end{figure}

\begin{figure}[!htbp]
\begin{center}
\safefig{\linewidth}{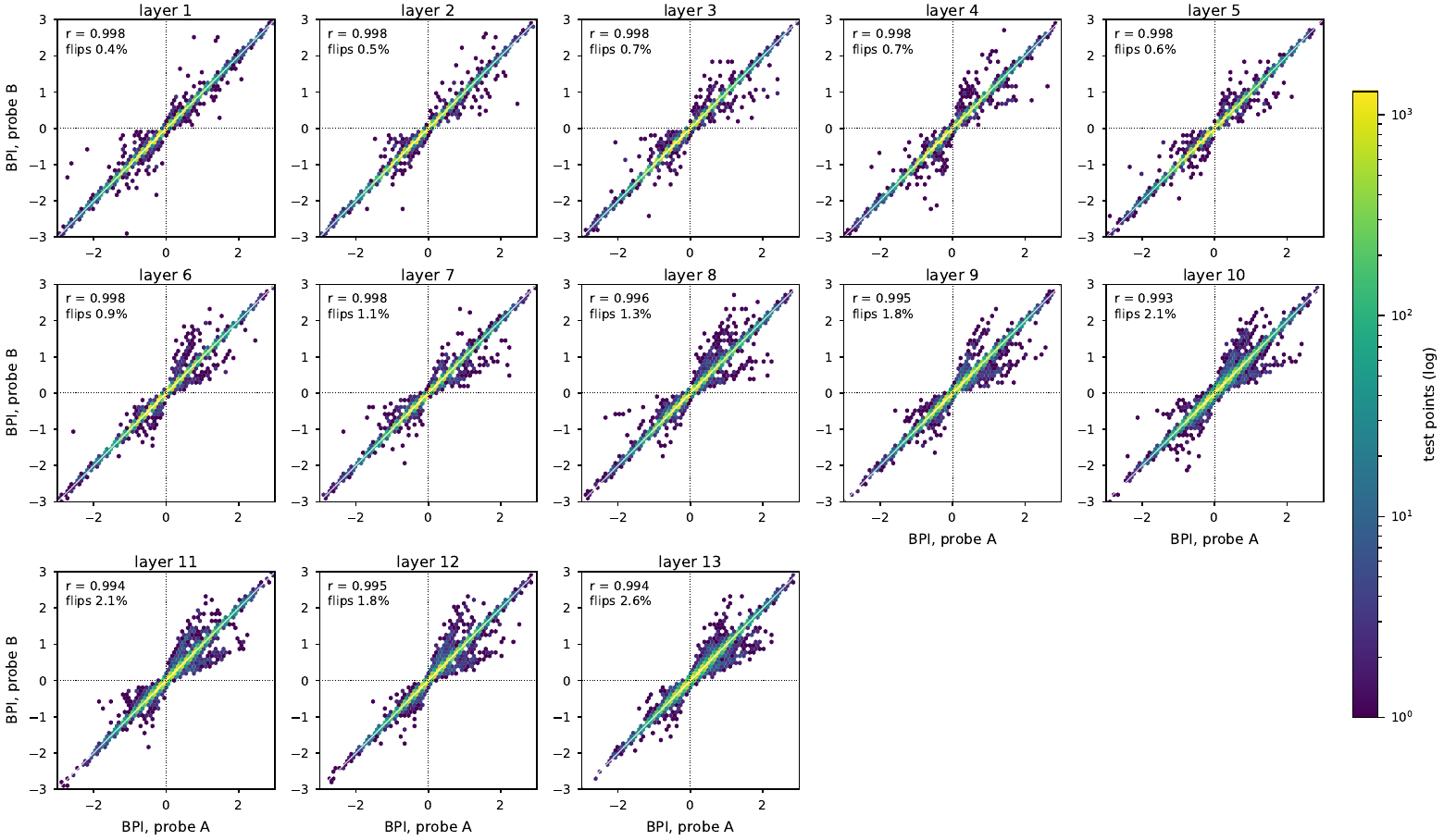}
\caption{As \cref{fig:n3_scff}, for VGG16 (the pretrained pipeline: CIFAR-10 upsampled to $224\times224$).}
\label{fig:n3_vgg16pre}
\end{center}
\end{figure}

Null~1 fits a second linear probe (seed 43) on the same untrained features as the paper's probe (seed 42), so a rescue by the second probe is probe noise. \Cref{fig:n3_scff,fig:n3_trifecta,fig:n3_asge,fig:n3_deeperforward,fig:n3_hff,fig:n3_hclff,fig:n3_monoforward,fig:n3_vgg16pre} compare the two probes for every method and every layer: BPI under the first probe against BPI under the second, for the test datapoints of the three untrained networks pooled. The probes are close to one decision function. The Pearson correlation of the two BPIs is 0.989 to 0.998 in every layer of every method; the classification of 0.4 to 3.1\% of the test datapoints differs between the probes (more in deeper layers), and these datapoints lie next to the boundary: under the first probe their median $|\mathrm{BPI}|$ is below 0.03 in every layer and the largest is 0.34. We compare what the probes do on the test set rather than their weights, because the features have thousands of dimensions and a low effective rank, so weight directions that the data barely constrain are set by the initialization and the weight decay while the predictions coincide.

\section{Other Datasets}
\label{app:datasets}

\Cref{fig:all_c10,fig:all_c100,fig:all_tin,fig:all_in1k} repeat \cref{fig:rate} with a row for every layer of the deepest network, on CIFAR-10, CIFAR-100, Tiny ImageNet and ImageNet-1K, and \cref{fig:f2l_c100,fig:f2l_tin} repeat \cref{fig:first_to_last} on CIFAR-100 and Tiny ImageNet; \cref{fig:net_rescue_c100,fig:net_rescue_tin} repeat \cref{fig:net_rescue}; the null controls on CIFAR-100 and Tiny ImageNet are in Appendix~\ref{app:null_ds}. The qualitative dependence is unchanged: on both datasets the rescue rate rises with BPI in every method, up to sampling noise, and rescue by a single layer of datapoints with $\mathrm{BPI}<-1$ is close to zero (at most 3\% pooled over transitions on CIFAR-100 and about 1\% or less on Tiny ImageNet). On ImageNet-1K (\cref{fig:all_in1k}) only three networks are shown: HFF-VGG16, which we trained for 30 epochs (7.7\% top-1 by the method's own read-out, against the 25.7\% its authors report; its curves characterize this trained instance, not HFF as published), the ImageNet-pretrained VGG16, and the untrained VGG16 as the three nulls of \cref{sec:null}; all three are read at 16 layers (13 convolutional, two 4096-wide fully connected, and the last layer: the class layer of HFF-VGG16, the logits of VGG16); probes there are fitted on 100 training images per class. On Tiny ImageNet and ImageNet-1K the raw-pixel probe is close to chance (3.7\% and 0.6\% test accuracy), so the row $0{\to}1$ is nearly uninformative there.

\section{Probes on Individual Layers}
\label{app:noncat}

\begin{figure}[t]
\begin{center}
\safefig{\linewidth}{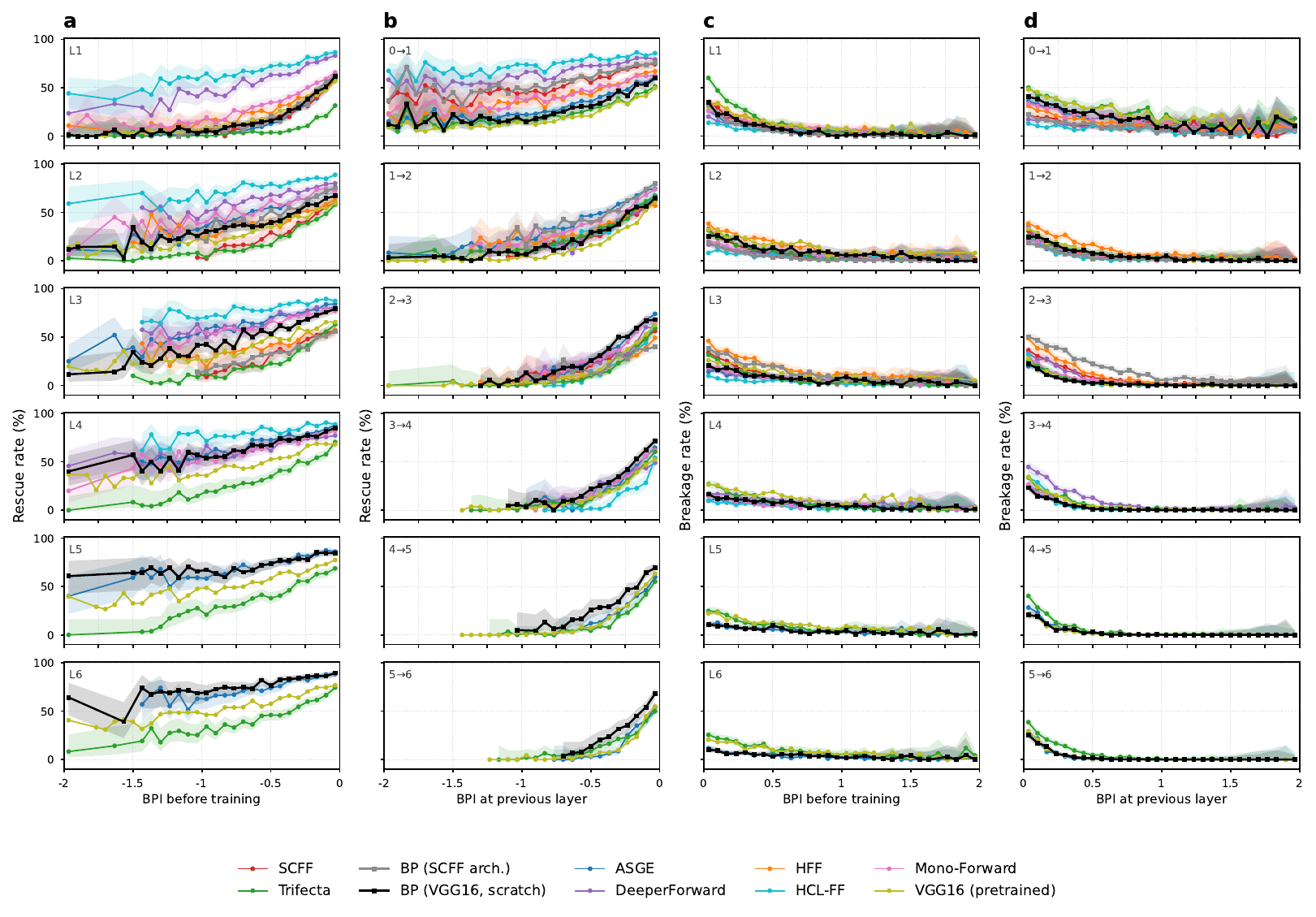}
\caption{As \cref{fig:rate}, on CIFAR-10, with every layer read on its own (no concatenation), for one seed and the first six layers.}
\label{fig:noncat_c10}
\end{center}
\end{figure}

\begin{figure}[t]
\begin{center}
\safefig{\linewidth}{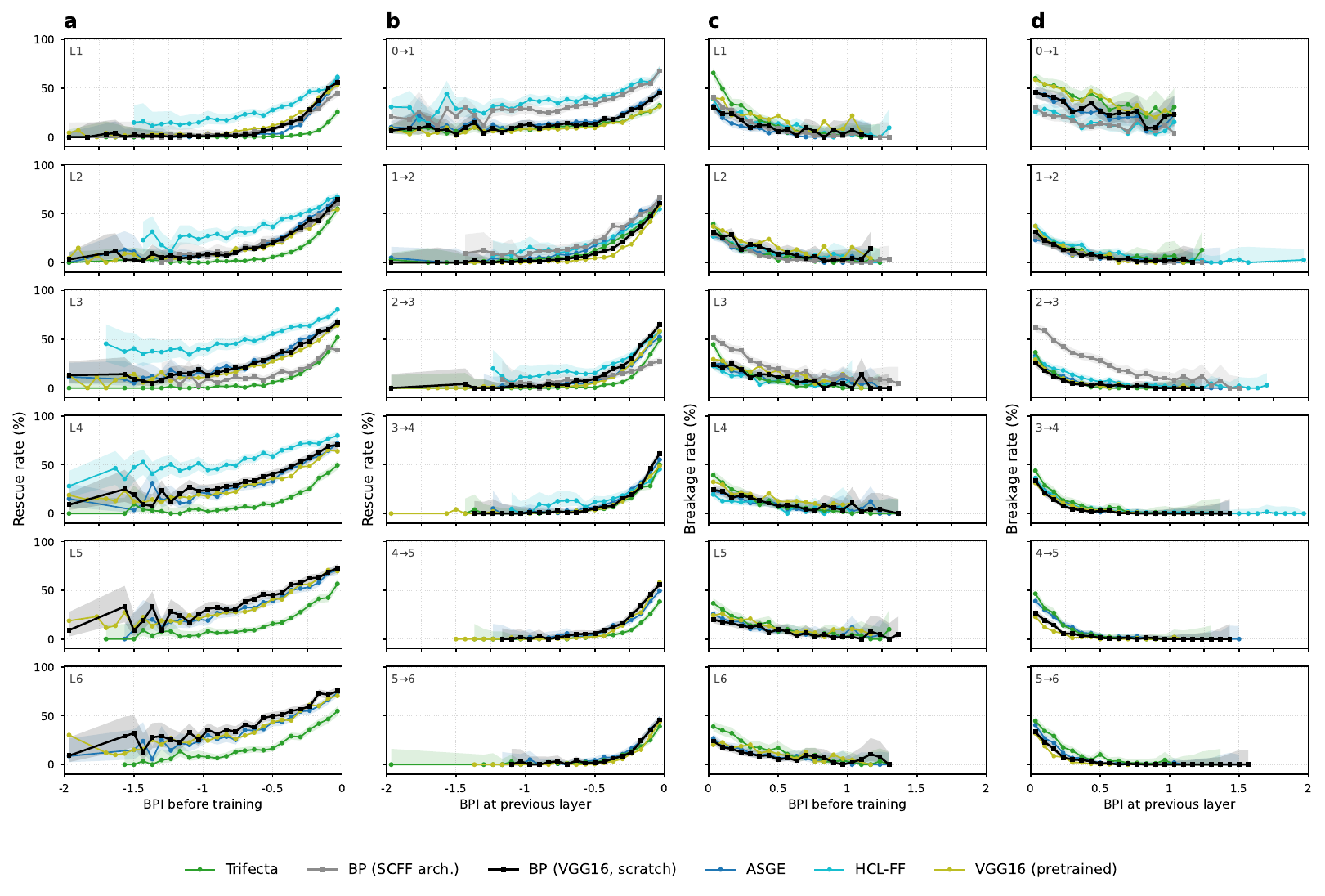}
\caption{As \cref{fig:noncat_c10}, on CIFAR-100 (without SCFF and DeeperForward, whose checkpoints are not available).}
\label{fig:noncat_c100}
\end{center}
\end{figure}

\begin{figure}[t]
\begin{center}
\safefig{\linewidth}{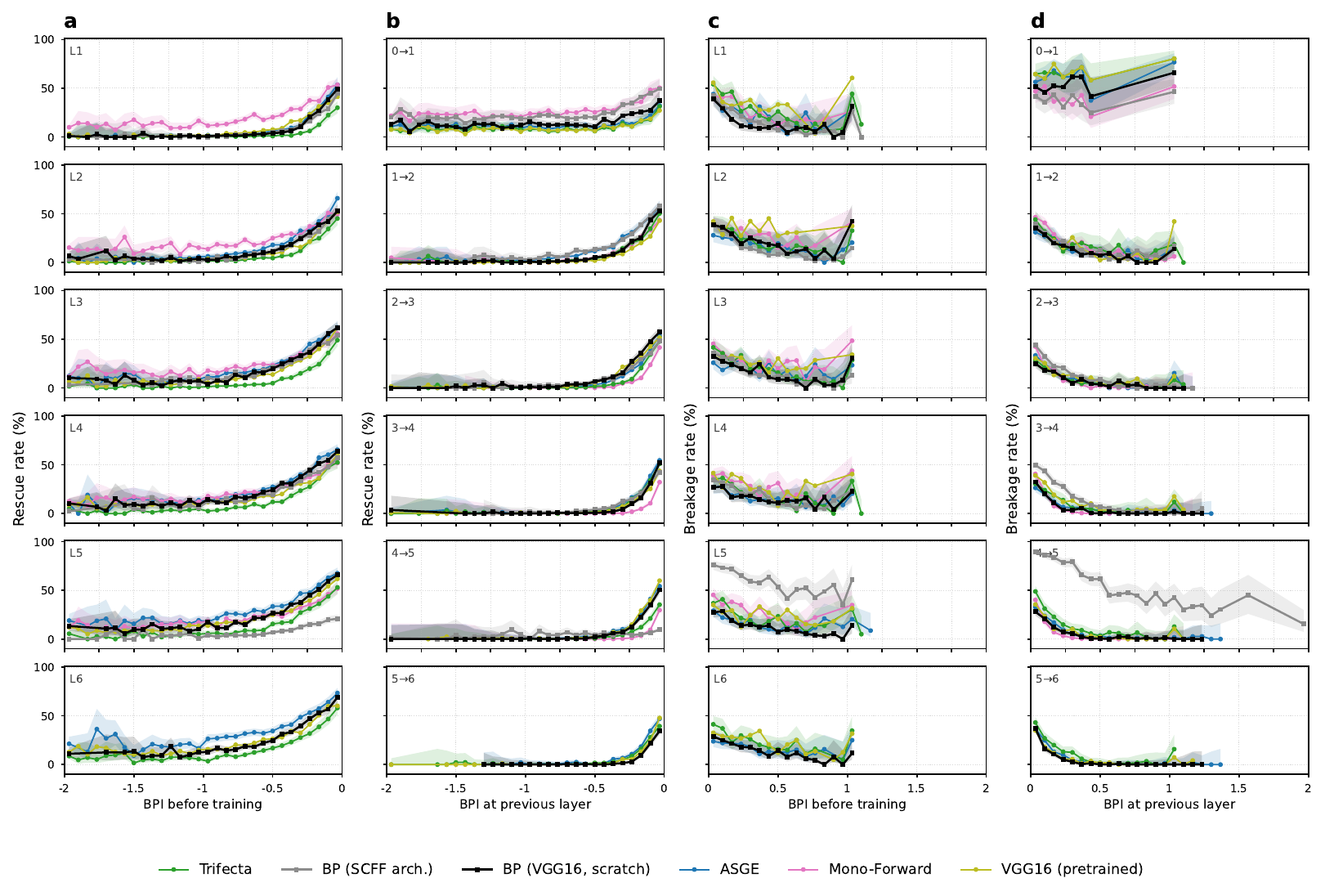}
\caption{As \cref{fig:noncat_c10}, on Tiny ImageNet (without SCFF and DeeperForward, whose checkpoints are not available).}
\label{fig:noncat_tin}
\end{center}
\end{figure}

\begin{figure}[t]
\begin{center}
\safefig{\linewidth}{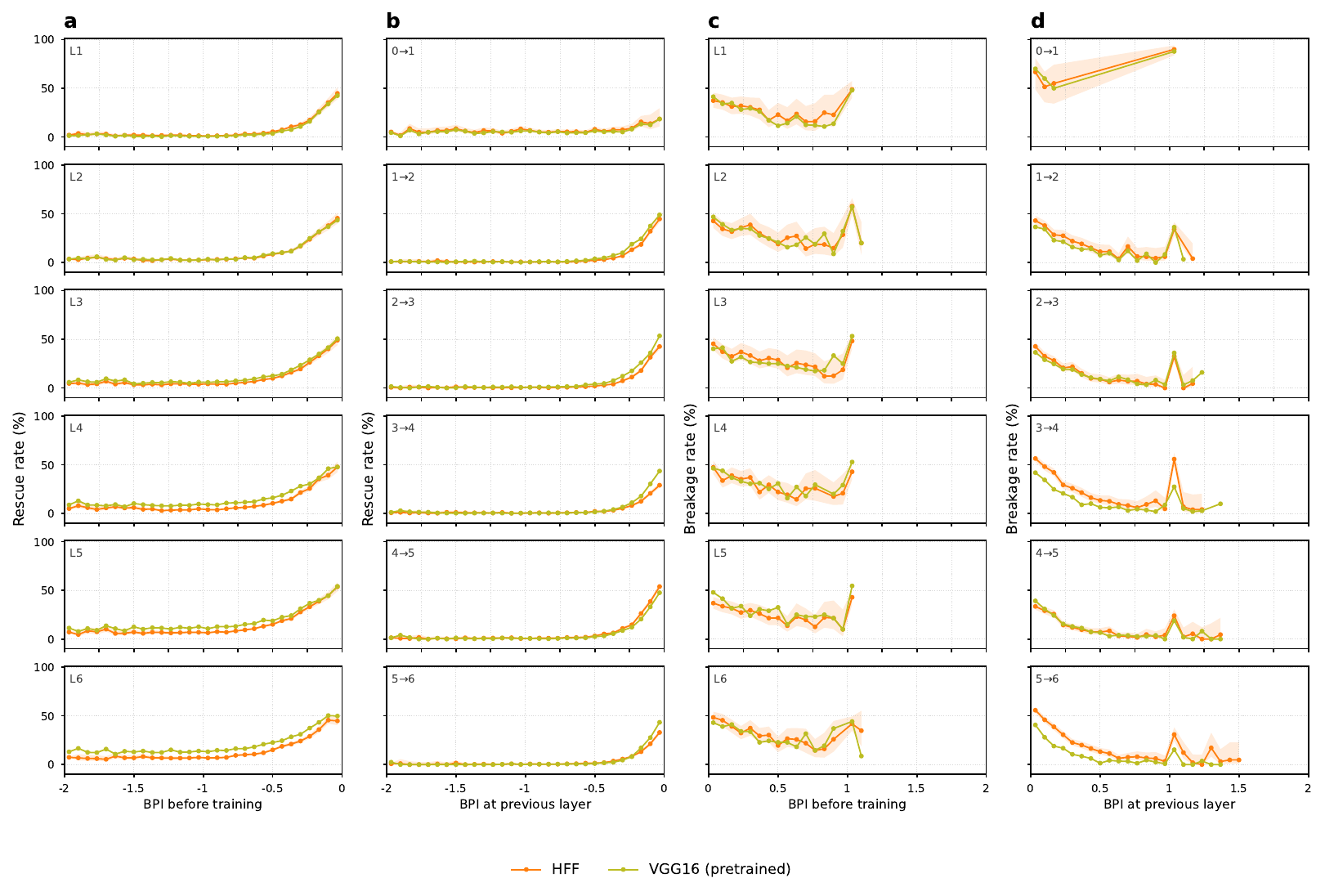}
\caption{As \cref{fig:noncat_c10}, on ImageNet-1K: HFF-VGG16 (our training) and the pretrained VGG16, against their untrained networks (probes fitted on 100 training images per class).}
\label{fig:noncat_in1k}
\end{center}
\end{figure}

The representation at depth $\ell$ in the main text is the concatenation of layers $1,\dots,\ell$ (Section~\ref{sec:untrained} and Appendix~\ref{app:setup}), which is not the read-out of every method. To check that our conclusions do not depend on this choice, we recomputed the rescue and breakage curves of \cref{fig:rate} with a linear probe that reads every layer on its own, without concatenation, for the untrained and the trained network. It covers the first six layers of every network and a single seed (seed~1), so its bands are wider than in the main text (fewer datapoints are pooled) and its numbers are not the seed-pooled ones of the main text (\cref{fig:noncat_c10,fig:noncat_c100,fig:noncat_tin,fig:noncat_in1k}). Layer~1 has nothing to concatenate, so its probe is the probe of the main text: its correctness agrees with the stored one on 96.5 to 100\% of the test datapoints in all 24 networks, and every other step (probe fitting, BPI, binning, the raw-pixel probe of row $0{\to}1$) is unchanged. SCFF and DeeperForward are missing on CIFAR-100 and Tiny ImageNet because their checkpoints are not available to us (only their stored arrays are), whereas BP on SCFF's architecture, which has no across-training curve in the main-text figures for these two datasets, has one here.

The qualitative dependence of rescue and breakage on BPI persists on all four datasets. In every network and layer, the rescue rate rises toward the boundary, both across training and across depth, and breakage is confined to datapoints close to the boundary. The methods keep their order: at layer~3 on CIFAR-10, HCL-FF and DeeperForward rescue the most datapoints with $\mathrm{BPI}<-1$ and Trifecta and SCFF the fewest, with or without concatenation. The magnitude of the change depends on the method and comparison: on CIFAR-10 (seed~1), the share of datapoints with $\mathrm{BPI}<-1$ that training rescues at layer~3 is 70\% with single-layer probes against 62\% with concatenation for HCL-FF, 45\% against 41\% for ASGE and 7.6\% against 2.9\% for SCFF, and the share of those within $0.5$ of the boundary changes by at most 4 points for all networks except BP on SCFF's architecture. That network is the one clear change: read on its own, its third layer rescues 46\% of the datapoints within $0.5$ of the boundary, against 65\% with concatenation, and on CIFAR-100 and Tiny ImageNet its last layer alone is a much worse read-out than the earlier layers (45.1\% against 58.6\% at the second layer on CIFAR-100; 15.0\% against 38.6\% at the fourth layer on Tiny ImageNet, below its untrained 21.2\%), which the concatenated representation hides. A single layer still rarely rescues datapoints far from the boundary (\cref{sec:across_depth}): at the transition $1{\to}2$ on CIFAR-10, at most 8\% of the datapoints with $\mathrm{BPI}<-1$ for DeeperForward, HCL-FF, SCFF, Trifecta and both VGG16 networks, and 14 to 25\% for ASGE, Mono-Forward, HFF and BP on SCFF's architecture (with concatenation, seed~1: at most 8\% for the first group and 10 to 15\% for the second).

\end{document}

%% file: null_table_rows.tex
SCFF & 7.1/0.9/0.0 & 50.7/33.6/4.6 & 0.7/0.7/0.0 & 44.4/37.2/2.8 & 43.9/34.6 \\
Trifecta & 2.3/0.0/0.0 & 38.3/20.9/4.6 & 7.7/0.2/0.0 & 50.1/24.2/3.0 & 58.2/34.6 \\
DeeperForward & 44.3/5.7/0.0 & 74.9/36.1/3.2 & 1.1/1.5/0.0 & 48.5/37.3/2.5 & 50.2/32.6 \\
HCL-FF & 59.8/5.7/0.0 & 84.5/35.7/3.3 & 5.3/1.1/0.0 & 50.5/37.1/2.5 & 52.7/34.9 \\
HFF & 22.1/1.9/0.0 & 47.5/32.0/4.4 & 7.2/0.9/0.0 & 46.5/34.4/3.0 & 37.3/31.7 \\
Mono-Forward & 37.0/0.6/0.0 & 72.3/27.5/4.4 & 14.6/1.5/0.0 & 60.1/36.8/3.7 & 66.7/42.7 \\
ASGE & 39.9/0.0/0.0 & 75.5/15.9/4.3 & 15.1/0.4/0.0 & 63.6/26.1/3.0 & 78.2/41.2 \\
BP (SCFF arch.) & 15.4/0.9/0.0 & 64.3/33.6/4.6 & 13.1/0.7/0.0 & 62.2/37.2/2.8 & 55.9/34.6

%% file: null_table_rows_cifar100.tex
BP (VGG16, scratch) & 10.4/0.1/0.0 & 48.7/6.0/0.8 & 0.6/0.1/0.0 & 32.2/10.2/0.8 & 58.9/27.2 \\
VGG16 (pretrained) & 7.8/0.1/0.0 & 43.6/6.0/0.8 & 0.1/0.1/0.0 & 24.1/10.2/0.8 & 65.7/27.2 \\
SCFF & 2.3/0.2/0.0 & 31.0/20.0/3.7 & 0.3/0.3/0.0 & 29.8/23.4/2.4 & 26.9/19.3 \\
Trifecta & 0.4/0.0/0.0 & 21.8/10.7/1.5 & 2.5/0.0/0.0 & 33.0/16.5/1.2 & 31.7/21.2 \\
DeeperForward & 18.9/3.3/0.0 & 54.8/22.9/2.7 & 1.4/0.8/0.0 & 33.2/27.0/2.0 & 29.8/16.4 \\
HCL-FF & 29.8/4.7/0.0 & 62.8/23.2/2.8 & 3.8/1.0/0.0 & 37.4/23.9/2.3 & 45.7/14.7 \\
ASGE & 10.7/0.0/0.0 & 49.2/8.4/1.4 & 3.5/0.1/0.0 & 41.5/18.9/1.0 & 50.7/25.2

%% file: null_table_rows_tinyimagenet.tex
BP (VGG16, scratch) & 4.2/0.1/0.0 & 34.8/5.4/0.5 & 0.4/0.1/0.0 & 21.6/8.3/0.6 & 43.7/13.4 \\
VGG16 (pretrained) & 6.3/0.1/0.0 & 30.8/5.4/0.5 & 0.2/0.1/0.0 & 14.7/8.3/0.6 & 60.1/13.4 \\
SCFF & 1.0/0.4/0.0 & 21.1/13.7/2.7 & 0.3/0.2/0.0 & 19.1/15.5/1.8 & 16.0/12.3 \\
Trifecta & 0.9/0.1/0.0 & 18.8/6.1/0.7 & 1.1/0.1/0.0 & 19.6/9.1/0.7 & 20.7/9.0 \\
DeeperForward & 10.5/1.1/0.0 & 38.7/17.4/2.2 & 0.9/0.8/0.0 & 23.6/20.3/1.1 & 18.0/10.9 \\
Mono-Forward & 14.3/0.9/0.0 & 34.8/9.4/1.6 & 0.3/0.1/0.0 & 19.5/7.8/1.2 & 13.0/6.1 \\
ASGE & 8.2/0.0/0.0 & 37.5/4.9/0.7 & 2.3/0.4/0.0 & 29.8/11.8/0.8 & 39.4/14.4